%% file: main.tex
\documentclass{article}

\usepackage{microtype}
\usepackage{graphicx}
\usepackage{booktabs} 
\usepackage{subcaption}

\usepackage{hyperref}
\usepackage{empheq}
\usepackage{xr}
\usepackage{mathdots}
\usepackage[OT1]{fontenc}
\usepackage[utf8]{inputenc}
\usepackage{multirow}
\usepackage[table,figure]{xcolor}
\usepackage{color}
\usepackage{tcolorbox}
\tcbuselibrary{listings,skins}
\usepackage{framed}
\newcommand{\mb}[1]{\mathbf{#1}}
\newcommand{\mc}[1]{\mathcal{#1}}
\definecolor{light-gray}{gray}{0.95}

\newcommand{\textbt}[1]{\textbf{\textit{#1}}}
\newcommand{\inv}{^{\raisebox{.2ex}{$\scriptscriptstyle-1$}}}
\newcommand\norm[1]{\left\lVert#1\right\rVert}
\newcommand\normx[1]{\left\Vert#1\right\Vert}
\newenvironment{proofs}{\paragraph{\normalfont \textit{Proof Sketch}.}}{\hfill$\square$\vspace{2mm}}

\usepackage[accepted]{icml2023-diffxyz}

\usepackage{amsmath}
\usepackage{amssymb}
\usepackage{mathtools}
\usepackage{amsthm}
\usepackage{csvsimple}
\usepackage{siunitx}

\usepackage[capitalize,noabbrev]{cleveref}

\theoremstyle{plain}
\newtheorem{theorem}{Theorem}[section]
\newtheorem{proposition}[theorem]{Proposition}

\theoremstyle{definition}

\theoremstyle{remark}

\usepackage[textsize=tiny]{todonotes}

\icmltitlerunning{PMaF: Deep Declarative Layers for Principal Matrix Features}

\begin{document}

\twocolumn[
\icmltitle{PMaF: Deep Declarative Layers for Principal Matrix Features}

\icmlsetsymbol{note}{*}

\begin{icmlauthorlist}
\icmlauthor{Zhiwei Xu}{anu}
\icmlauthor{Hao Wang$^\dag$}{anu}
\icmlauthor{Yanbin Liu}{anu}
\icmlauthor{Stephen Gould}{anu}
\end{icmlauthorlist}

\icmlaffiliation{anu}{School of Computing, CECC, ANU, Canberra, Australia}

\icmlcorrespondingauthor{Zhiwei Xu}{zhiwei.xu@anu.edu.au}

\icmlkeywords{Machine Learning, ICML}

\vskip 0.3in
]

\printAffiliationsAndNotice{$^\dag$Parts of this work were completed when Hao Wang pursued a master's degree at the Australian National University. $^\ddag$PMaF for the IED problem is equivalent to principal component analysis \cite{pca} for a covariance matrix.}

\begin{abstract}
We explore two differentiable deep declarative layers, namely least squares on sphere (LESS) and implicit eigen decomposition (IED), for learning principal matrix features (PMaF). It can be used to represent data features with a low-dimensional vector containing dominant information from a high-dimensional matrix. We first solve the problems with iterative optimization in the forward pass and then backpropagate the solution for implicit gradients under a bi-level optimization framework. Particularly, adaptive descent steps with the backtracking line search method and descent decay in the tangent space are studied to improve the forward pass efficiency of LESS. Meanwhile, exploited data structures are used to greatly reduce the computational complexity in the backward pass of LESS and IED. Empirically, we demonstrate the superiority of our layers over the off-the-shelf baselines by comparing the solution optimality and computational requirements.
\end{abstract}

\section{Introduction}
Principal matrix feature (PMaF)$^\ddag$ in this work refers to a single vector summarising a data matrix.
It can be used in deep feature representation or learning that is typical in various areas, such as image analysis \cite{feature-learn-medical,vgg_spectral}, natural language processing \cite{feature-learn-language}, weather prediction \cite{feature-learn-weather}, and so on.
It adapts learned features for downstream tasks and studies matrix structures for fine-grained features with such as a high sparsity or a low dimension \cite{sparse-feats,low-rank-feats,cons-deep-behaviour,cons-kernel}.

In this work, we mainly focus on two optimization problems for PMaF and study two deep layers, namely least squares on sphere (LESS) and implicit eigen decomposition (IED), that are superior to off-the-shelf SciPy (nondifferentiable) \cite{scipy} and PyTorch \cite{pytorch} baselines in the effectiveness and/or efficiency of optimization and differentiability for end-to-end learning.

\begin{figure}[t]
\centering
\includegraphics[width=0.47\textwidth]{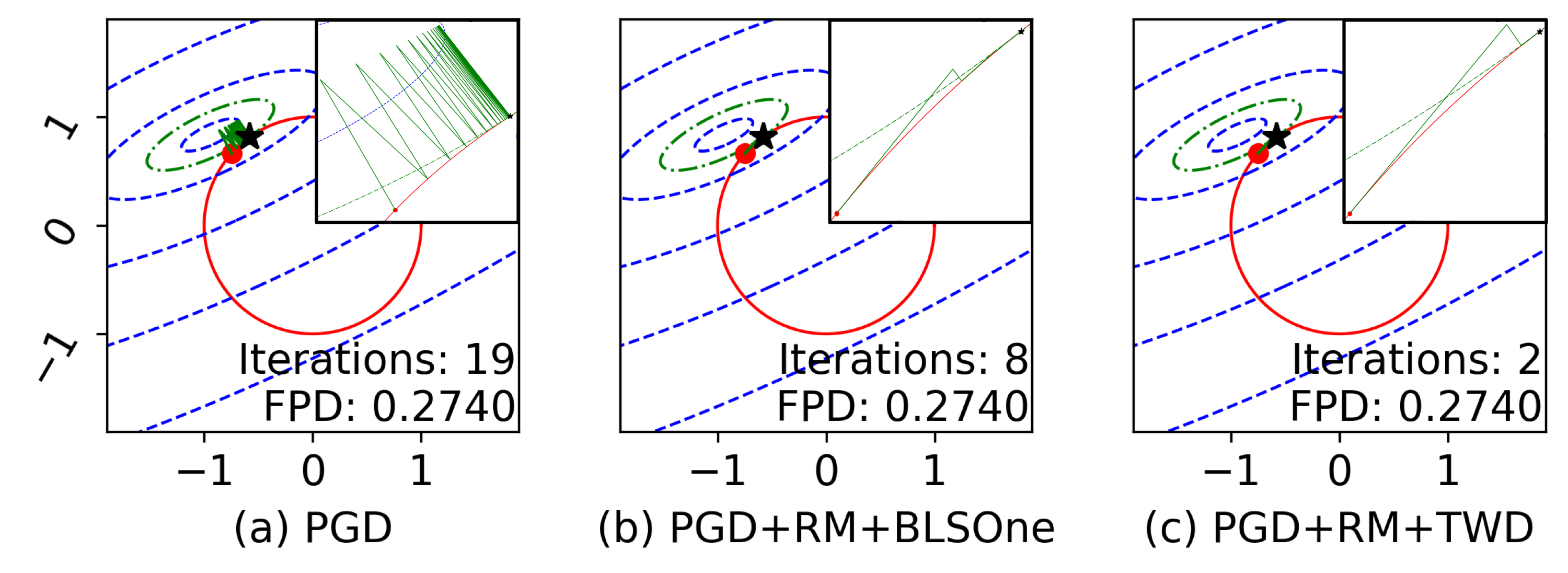}
\vspace{-5mm}
\caption{\textit{With BLS and TWD, LESS converges faster and better.
The moving path starts from the initial value ({\color{red}\textbf{red}}-dot) to the optimal (\textbf{black}-star).
More results are in the Appendix.}}
\vspace{-4mm}
\label{fig:case_study_short}
\end{figure}

In LESS, the proposed adaptive gradient descent steps on the tangent plane greatly reduce the number of iterations, see Fig.~\ref{fig:case_study_short}; in IED, the alternatives, power iteration (PI) \cite{power_iteration} and simultaneous iteration (SI) \cite{s_iteration}, achieve better solutions for non-negative symmetric and nonsymmetric matrices than the baseline.
Meanwhile, we use implicit differentiation methods, mainly deep declarative network (DDN) \cite{ddn} in this work, with exploited matrix structures \cite{exploit} to dramatically reduce the computational requirements.
Comprehensive experiments are provided in the Appendix.

\section{Deep Declarative Layers}

\subsection{Least Squares on Sphere}

Given $\mb{A} \in \mathbb{R}^{m \times n}$ and $\mb{b} \in \mathbb{R}^{m}$, the least squares problem with solution constrained on a unit sphere is defined as
\begin{equation}
\begin{gathered}
\label{eq:obj}
\text{minimize}_{\mb{u} \in \mathbb{R}^n} \texttt{f}(\mb{A},\mb{b}, \mb{u}) \triangleq \frac{1}{2} \norm{ \mb{A}\mb{u}-\mb{b} }^2 \ ,\\
\text{subject to}\  \norm{ \mb{u} }^2=1\ ,
\end{gathered}
\end{equation}
where $\norm{\cdot}$ is the $\ell_2$-norm.
For the notation simplicity, we denote $\texttt{f}(\mb{A},\mb{b}, \mb{u})$ by using $\texttt{f}(\mb{u})$.
Since $\mb{u}$ is constrained on the sphere, the closed-form of the optimal $\mb{y}=\mb{A}\inv\mb{b}$ no longer holds.
The projected gradient descent (PGD) method is used to decrease the energy in the gradient descent direction while guaranteeing the solution feasibility.

We first describe the vanilla PGD, PGD with direction weight decay, and PGD projected onto the Riemannian manifold (directly on the constraint sphere).
For monotonic energy convergence, either the backtracking line search method or a simple yet effective step decay in the tangent space is used for fast optimal solution search.

\textbf{1) \textit{Projected Gradient Descent (PGD).}}
We refer to the Appendix for the details of the widely used PGD method \cite{pgd_1,pgd_2}.

\textbf{2) \textit{Direction Weight (DW).}}
Since the descent step $\eta$ in PGD needs to decrease for a fine search when $\mb{u}_t$ approaches the optimal solution, reducing $\eta$ with weight $w_t$ adaptive to the descent direction is desirable.
Given $\texttt{f}(\mb{u})$ in Eq.~\eqref{eq:obj}, the center $\mb{u}_0=(\mb{A}^\top\mb{A})\inv\mb{A}^\top\mb{b}$ indicates if the least squares problem is an inner (non-convex) or outer (convex) equality-constrained problem and also the direction from the constraint center to the solution.

With the optimal unconstrained solution, $\mb{u}_0$ is parallel to the descent direction $\mb{d}_t$.
Hence, we define a direction weight for each descent step as
\begin{equation}
\label{eq:dir_weight}
w_t = 1 - \texttt{S}_\texttt{c}(\mb{d}_t, \mb{u}_0)\ ,
\end{equation}
where
\begin{equation}
\label{eq:dir_weight_dir}
\begin{aligned}
\mb{d}_t=
\begin{cases}
    -\nabla \texttt{f}(\mb{u}_t) & \text{if $\norm{ \mb{u}_0 } \geq 1$} \\
     \nabla \texttt{f}(\mb{u}_t) & \text{otherwise}
\end{cases}
\end{aligned}
\end{equation}
and $\texttt{S}_\texttt{c}(\mb{a},\mb{b})=\mb{a}^\top \mb{b}/\left(\norm{ \mb{a} } \norm{ \mb{b} }\right)$ is the cosine similarity to measure the direction homotropy.
When $\mb{u}_t$ is optimal, $w_t=0$ such that $w_t \eta=0$ terminates the update of $\mb{u}_t$.

\textbf{3) \textit{PGD on Riemannian Manifold (RM).}}
As the solution is constrained on a sphere, using Riemannian geodesic distance is promising to learn better feature distribution than using the Euclidean distance \citep{rieman-geodesic-brain,rieman-geodesic-domain}.
Therefore, we define the solution projection onto the Riemannian manifold \citep{boumal2020intromanifolds} as
\begin{equation}
\begin{aligned}
\label{eq:rie}
\texttt{Proj}_{\texttt{RM}}\left( -\nabla \texttt{f}(\mb{u}_t) \right) = \left( \mb{I}_n - \mb{u}_t \mb{u}^\top_t \right) \left( -\nabla \texttt{f}(\mb{u}_t) \right)\ ,
\end{aligned}
\end{equation}
where $\mb{I}_n$ is an $n \times n$ identity matrix.
The solution update at $(t+1)$ follows
\begin{equation}
\begin{aligned}
\label{eq:rie_solution_update}
\mb{u}_{t+1} = \texttt{Proj}_{\texttt{Sph}}\left( \mb{u}_t + \eta \texttt{Proj}_{\texttt{RM}} \left( -\nabla \texttt{f}(\mb{u}_t) \right) \right)\ ,
\end{aligned}
\end{equation}
where $\eta=1$ or can be derived from $\nabla \texttt{f}(\mb{u}_{t+1}) = 0$ as
\begin{equation}
\begin{aligned}
\label{eq:rie_eta}
\eta = \frac{\left( \mb{A}\mb{u}_t - \mb{b} \right)^\top \mb{A} \left( \mb{I}_n - \mb{u}_t \mb{u}^\top_t \right) \mb{A}^\top (\mb{A} \mb{u}_t - \mb{b})}{\norm{ \mb{A}(\mb{I}_n - \mb{u}_t \mb{u}^\top_t) \mb{A}^\top (\mb{A} \mb{u}_t - \mb{b}) }^2}\ .
\end{aligned}
\end{equation}

See the Appendix for the Riemannian manifold projection.

\textbf{4) \textit{Backtracking Line Search (BLS).}}
While the direction weight method causes many iterations if the solution is far from optimal, a guarantee of energy reduction in every iteration is crucial, requiring a suitable descent step.
For this to hold, we refer to the backtracking line search method \citep{boyd2004convex} and apply the first-order Lagrangian form of Eq.~\eqref{eq:obj}
\begin{equation}
\begin{aligned}
\label{eq:backtrack}
\texttt{f}(\mb{u}_t+\eta \Delta \mb{u}_t) \leq \texttt{f}(\mb{u}_t) + \alpha \eta \nabla^\top \texttt{f}(\mb{u}_t) \Delta \mb{u}_t\ ,
\end{aligned}
\end{equation}
where constant $\alpha \in (0, 0.5)$ is for the maximum energy decrease at the $(t+1)^{\text{th}}$ iteration for the descent monotonicity.
Otherwise, the descent step would be decayed by $\eta \leftarrow \beta \eta$ with $\beta \in (0, 1)$ to avoid surpassing the optimal solution.

\textbf{5) \textit{Tangent Weight Decay (TWD).}}
Alternatively, since a large descent step tackles the monotonic energy decrease in the update around the optimal solution, it could be unable to converge on the manifold.
Hence, the descent step is decayed in the tangent space on the same side of the descent direction with decay rate $\beta$, that is $\eta \leftarrow \beta \eta$, when
\begin{equation}
\begin{aligned}
\label{eq:tandecay}
\texttt{S}_\texttt{c} \left( \texttt{Proj}_{\texttt{RM}} \left( -\nabla \texttt{f}(\mb{u}_t) \right), \texttt{Proj}_{\texttt{RM}} \left( -\nabla \texttt{f}(\mb{u}_{t+1}) \right) \right) < 0\ ,
\end{aligned}
\end{equation}
with
\begin{equation}
\begin{aligned}
\label{eq:tandecay_update}
\mb{u}_{t+1} = \mb{u}_{t} + \eta \texttt{Proj}_{\texttt{RM}} \left( -\nabla \texttt{f}(\mb{u}_t) \right)\ .
\end{aligned}
\end{equation}
In short, Eq.~\eqref{eq:tandecay} indicates that the solution update at the $(t+1)^{\text{th}}$ iteration causes a ``reverse" descent direction, usually in $(90, 180]$ degrees, around the optimal solution, and thus, a smaller step is preferred.

\subsection{Implicit Eigen Decomposition}

Generally, the eigen decomposition problem can be formulated by solving the optimization $\texttt{f}: \mb{A} \in \mathbb{R}^{m \times m} \rightarrow \{ \boldsymbol{\lambda}, \mb{u} \}$ with $\boldsymbol{\lambda}$ as a vector of $n$ largest eigenvalues and $\mb{u} \in \mathbb{R}^{m \times n}$ as the corresponding eigenvectors,
\begin{equation}
\begin{gathered}
\label{eq:ed}
\text{minimize}_{\mb{u} \in \mathbb{R}^{m \times n}} \texttt{f}(\mb{A}, \mb{u}) \triangleq -\texttt{tr} \left( \mb{u}^\top \mb{A} \mb{u} \right)\ ,\\
\text{subject to} \quad \texttt{h} \left( \mb{u} \right) \triangleq \mb{u}^\top \mb{u} = \mb{I}_n\ ,
\end{gathered}
\end{equation}
where \texttt{tr}() is the trace function over all $n$ eigenvalues.
The optimal solution in Eq.~\eqref{eq:ed_argmin} satisfies Eq.~\eqref{eq:eigen_attribute} as
\begin{align}
\mb{y} &= \texttt{argmin}_{\mb{u} \in \mathbb{R}^{m \times n}} \texttt{f}(\mb{A}, \mb{u})\label{eq:ed_argmin}\ ,\\
\mb{A} \mb{y}_i &= \lambda_i \mb{y}_i, \quad \forall i \in \mathcal{N} = \{1, ..., n\} \label{eq:eigen_attribute}\ .
\end{align}
The principal matrix component refers to the eigenvector associated with the largest
eigenvalue, for which we use the power iteration algorithm and the simultaneous iteration algorithm.
For a complete analysis, solution update formulas are provided below.

\textbf{Solver 1. \textit{Power Iteration (PI)}.}
Given a randomly initialized eigenvector $\mb{u}_0$, the solution update at
time $(t+1)$ follows
\begin{equation}
\label{eq:pi_update}
    \mb{u}_{t+1} = \mb{A} \mb{u}_t / \norm{ \mb{A} \mb{u}_t }\ ,
\end{equation}
and terminates (upon the convergence or the maximum iteration) at $t=K$ for the principal eigenvector and eigenvalue,
\begin{equation}
\label{eq:pi_optimal}
    \mb{y} = \mb{u}_K\ \quad \text{and} \quad
    \lambda = \mb{y}^\top \mb{A} \mb{y}\ .
\end{equation}
%
\textbf{Solver 2. \textit{Simultaneous Iteration (SI)}.}
QR decomposition is required for iterative updates of the input and the solution.
The initial input $\mb{x}_0=\mb{A}$ and the updates at time $t$ follow
\begin{equation}
\label{eq:si_update}
    \{\mb{Q}_t, \mb{R}_t\} = \texttt{QR}(\mb{x}_t) \quad \text{and} \quad
    \mb{x}_{t+1} = \mb{x}_t \mb{Q}_t\ .
\end{equation}
The principal eigenvector is the component of $\mb{Q}_K$ corresponding to the largest eigenvalue $\lambda=\texttt{max} (\mb{R}_K)$.

\textbf{\textit{Solution consistency for effective backpropagation}.}
The eigen decomposition problem defined in Eq.~\eqref{eq:ed} has two optimal solutions with reverse directions due to the quadratics.
This could happen either in the intermediately updated solution $\mb{u}_t$ (either Eq.~\eqref{eq:pi_update} or Eq.~\eqref{eq:si_update}) or in the optimal solution $\mb{y}$ in each learning epoch for the same data sample.
To alleviate the ineffectiveness of these updates, one can apply a reference direction $\mb{r}$
such that the optimal solution $\mb{y}$ updates in the same direction of $\mb{r}$ as
\begin{align}
    \textit{Historical}: \quad \mb{u}_t &\leftarrow \texttt{V}\left( \mb{u}_t, \mb{u}_{t-1} \right) \mb{u}_t\ ,\\
    \textit{Hard-coded}: \quad \mb{y} &\leftarrow \texttt{V}\left( \mb{y}, \mb{r} \right) \mb{y}\ ,
\end{align}
where $\texttt{V}(\mb{a}, \mb{b})=\texttt{Sign} \left( \mb{a}^\top \mb{b} \right)$ if $(\mb{a} \not\perp \mb{b})$ and otherwise $1$ and $\texttt{Sign}()$ calculates the sign value of a scalar. If $\mb{u}_t$ is for eigenvectors of multiple eigenvalues, only the diagonals of $\mb{u}_t^\top \mb{u}_{t-1}$ are used for the sign values.
\vspace{-2mm}

\section{Implicit Differentiation}
Given the optimal solutions, their gradients over the input entries enable end-to-end learning.
Without unrolling the iterations of the solvers, we use a single-step method with implicit differentiation \cite{ddn}, fixed-point theorem (for IED), and exploited structures \cite{exploit}.

\subsection{Deep Declarative Networks based Gradients}
\label{sec:ddn}
With the iterative optimization as the forward pass, the backward pass of $\mb{y}$ to $\mb{A}$ in Eq.~\eqref{eq:obj} and Eq.~\eqref{eq:ed} is required.
Backtracking the forward pass, however, is inefficient and sometimes infeasible due to the discrete solution.
Hence, we use the deep declarative network method \cite{ddn} to efficiently calculate $\nabla_X L$ with $\nabla_X \mb{y}$, where $L$ is the loss from the upper problem in the bi-level optimization and $X$ and $Y$ indicate the input variable (can be multiple) and the optimization solution respectively.
It follows
\begin{align}
\mc{K} = \nabla_Y L &\left( \mc{H}\inv \mc{A}^\top \left( \mc{A} \mc{H}\inv \mc{A}^\top \right)\inv \mc{A} - \mb{I}_n \right) \mc{H}\inv \label{eq:k_comp} \ ,\\
&\qquad \quad \nabla_X L = \mc{K} \mc{B} \label{eq:backward_ddn}\ .
\end{align}

\textbf{1) \textit{LESS}.}
The components of $\mc{K}$ in Eq.~\eqref{eq:k_comp} are
\begin{align}
\mc{A} &= 2 \mb{y}^\top \in \mathbb{R}^{1 \times n}\ , \\
\label{eq:hessian_less}
\mc{B} &= \nabla^2_{XY} \texttt{f}(\mb{A}, \mb{b}, \mb{y}) \in \mathbb{R}^{n \times (m \times n)}\ , \\
\mc{H} &= \mb{A}^\top \mb{A} - 2 \beta \mb{I}_n \in \mathbb{R}^{n \times n}\ , \\
\beta &= \frac{1}{2} \mb{y}^\top \mb{A}^\top \left( \mb{A} \mb{y} - \mb{b} \right) \in \mathbb{R}\ .
\end{align}
\begin{proposition}
\label{prop:less_hessian}
    (Exploited Hessian structure for LESS) Rather than applying Jacobian and Hessian operations, an accumulation by parts approach with the exploited structure of $\mc{B}$ greatly improves the implementation efficiency.
    Recall that $\mc{B} \in \mathbb{R}^{n \times (m \times n)}$, $\mb{A} \in \mathbb{R}^{m \times n}$, and $\mb{y} \in \mathbb{R}^{n}$.
    Set indices $i,j \in \mc{N}=\{1, ..., n\}$ for the dimensions related to $n$ such that $\mc{B}=[\mc{B}_{ij}]$, $\mb{A}=[\mb{A}_i]$, and $\mb{y}=[\mb{y}_j]$. Then,
    \begin{empheq}[box=\fbox]{align}
        \mc{B}_{ij} &= \mb{A}_i \mb{y}^\top_j, \quad \forall i, j \in \mc{N}\ ,\\
        \mc{B}_{ii} &\leftarrow \mc{B}_{ii} + (\mb{A} \mb{y} - \mb{b}), \quad \forall i \in \mc{N}\ .
    \end{empheq}
\end{proposition}
\vspace{-3mm}
\begin{proofs}
See the Appendix for details.
\end{proofs}
\vspace{-3mm}

\textbf{2) \textit{IED}.} Similarly, for IED, one has
\begin{align}
\mc{A} &= 2 \mb{y}^\top \in \mathbb{R}^{1 \times m}\ , \\
\label{eq:hessian_ed}
\mc{B} &= \nabla^2_{XY} \texttt{f}(\mb{A}, \mb{y}) \in \mathbb{R}^{m \times (m \times m)}\ , \\
\mc{H} &= -\left( \mb{A} + \mb{A}^\top \right) - 2 \beta \mb{I}_m \in \mathbb{R}^{m \times m}\ , \\
\beta &= -\frac{1}{2} \mb{y}^\top \left( \mb{A} \mb{y} + \mb{A}^\top \mb{y} \right) \in \mathbb{R}\ .
\end{align}
\begin{proposition}
\label{prop:eigen_lagrange}
    If the input $\mb{A}$ of the objective function Eq.~\eqref{eq:ed} is symmetric, the Lagrange multiplier equals the negative largest eigenvalue, that is, $\beta=-\lambda$.
\end{proposition}
\vspace{-3mm}
\begin{proof}
Since $\nabla_Y \texttt{f}(\mb{A}, \mb{y})=\beta \nabla_Y \texttt{h}(\mb{A}, \mb{y})$ in Eq.~(17) of \cite{ddn} for the minima, $\nabla_Y \texttt{f}(\mb{A}, \mb{y})=-(\mb{A} \mb{y} + \mb{A}^\top \mb{y})^\top$ and $\nabla_Y \texttt{h}(\mb{A}, \mb{y})=2\mb{y}^\top$ for the problem defined in Eq.~\eqref{eq:ed}, and $\mb{A} \mb{y}=\lambda \mb{y}$, one has $-(\lambda \mb{y} + \mb{A}^\top \mb{y}) = 2 \beta \mb{y}$.
If $\mb{A}^\top = \mb{A}$, then $\mb{A}^\top \mb{y} = \mb{A} \mb{y} = \lambda \mb{y}$, and thus, $\beta = -\lambda$.
\end{proof}

\begin{proposition}
\label{prop:ied_hessian}
    (Exploited Hessian structure for IED)
    Recall that $\mc{B} \in \mathbb{R}^{m \times (m \times m)}$ and $\mb{y} \in \mathbb{R}^{m}$.
    Set indices $i,j,k \in \mathcal{M}=\{1, ..., m\}$ for the dimensions related to $m$ such that $\mc{B}=[\mc{B}_{ijk}]$ and $\mb{y}=[\mb{y}_i]$.
    Then,
    \begin{empheq}[box=\fbox]{align}
        \mc{B}_{ijk} &= 0, \quad \forall i,j,k \in \mc{M}\ ,\\
        \mc{B}_{iji} &\leftarrow \mc{B}_{iji} - \mb{y}_j, \quad \forall i,j \in \mc{M}\ ,\\
        \mc{B}_{iij} &\leftarrow \mc{B}_{iij} - \mb{y}_j, \quad \forall i,j \in \mc{M}\ .
    \end{empheq}
\end{proposition}
\vspace{-3mm}
\begin{proofs}
This can be easily proved by finding the structure of $\mc{B}$ from an example.
\end{proofs}
\vspace{-3mm}

Nevertheless, $\mc{B}$ requires ($4m^3/1024^2$)MB memory in the single-precision floating-point format, for instance, 
4,096MB for $m=1024$.
With the exploited Hessian structure of $\mc{B}$ and the vector-Jacobian product for $\nabla_X L$, however, $\mc{B}$ can be avoided by using merely $\mb{y}$.
\begin{proposition}
\label{prop:ied_gradient}
    (Exploited gradient structure for IED)
    $\mc{B}$ can be split into $\mb{y}$ dependencies for $\nabla_X L \in \mathbb{R}^{m \times m}$ as
    \begin{empheq}[box=\fbox]{align}
        \nabla_X L = -\mc{K} \mb{y}^\top - \mb{y} \mc{K}^\top\ .
    \end{empheq}
\end{proposition}
\vspace{-3mm}
\begin{proof}
See the Appendix for details.
\end{proof}
\vspace{-3mm}

\subsection{Implicit Function Theorem based Gradients}

The implicit function theorem based gradients only apply to eigen decomposition.
By using PI for the largest eigenvalue (to be positive), the implicit function is
\begin{equation}
\label{eq:ift_pi}
\texttt{f}(\mb{A}, \mb{u}_t, \mb{u}_{t+1}) = \mb{u}_{t+1} - \mb{A} \mb{u}_t / \norm{ \mb{A} \mb{u}_t }
\end{equation}
with the input matrix $\mb{A} \in \mathbb{R}^{m \times m}$ and the solution $\mb{u} \in \mathbb{R}^{m \times n}$.
Upon the convergence, $\mb{y}=\mb{u}_{t+1}=\mb{u}_t$, and thus,
\begin{equation}
\texttt{f}(\mb{A}, \mb{y}) = \mb{y} - \mb{A} \mb{y} / \norm{ \mb{A} \mb{y} }\ .
\end{equation}
Applying the implicit function theorem to $\texttt{f}(\mb{A}, \mb{y})$ achieves
\begin{align}
\nabla_X \texttt{f}(\mb{A}, \mb{y}) + \nabla_Y \texttt{f}(\mb{A}, \mb{y}) \nabla_X \mb{y} = 0\ , \\
\nabla_X \mb{y} = -\left( \nabla_Y \texttt{f}(\mb{A}, \mb{y}) \right)\inv \nabla_X \texttt{f}(\mb{A}, \mb{y})\ .
\end{align}
For the notation consistency with Sec.~\ref{sec:ddn}, we denote $\mathcal{H} = \nabla_Y \texttt{f}(\mb{A}, \mb{y}) \in \mathbb{R}^{m \times m}$, $\mathcal{B} = \nabla_X \texttt{f}(\mb{A}, \mb{y}) \in \mathbb{R}^{m \times (m \times m)}$, and $\nabla_X \mb{y} = -\mathcal{H}\inv \mathcal{B} \in \mathbb{R}^{m \times (m \times m)}$.
Note that $\nabla_X \mb{y}$ is in the same form of Eq.~\eqref{eq:backward_ddn}
as an unconstrained problem.
Then, $\nabla_Y L \in \mathbb{R}^{m}$ follows
\begin{equation}
\nabla_X L = \nabla_Y L \nabla_X \mb{y} = - \nabla_Y L \mathcal{H}\inv \mathcal{B}\ ,
\end{equation}
\begin{empheq}[box=\fbox]{align}
\mc{H}
&= \mb{I}_m - \left( \mb{I}_m -\mb{y} \mb{y}^\top \right) \mb{A} / \lambda\ ,\\
\mc{B}
&= -\left( \mb{I}_m - \mb{y} \mb{y}^\top \right) \mb{y}^\top / \lambda \ .
\end{empheq}
Different from using multiple eigenvalues for gradients \cite{eigenvalue_grad}, only the largest one is used.
Again, it is strongly encouraged to use the vector-Jacobian product to calculate $\nabla_X L$ over $\mc{B}$ for memory reduction.

\section{Experiments}
Experiments are on a 3090 GPU and PyTorch 1.12.0.
The code is at \href{https://github.com/anucvml/ddn.git}{https://github.com/anucvml/ddn.git}.
See the Appendix for \textbf{evaluation standards} including fixed-point distance (FPD) in Eq.~\eqref{eq:fpd}, eigen distance in Eq.~\eqref{eq:edist}, and mean relative error (MRE) in Eq.~\eqref{eq:mre}.
We highlight \colorbox{gray!50}{ours}.

\textbf{4.1. Evaluation on LESS.}
We show the \textbt{effectiveness} in Fig.~\ref{fig:case_study_short} and Table~\ref{tb:table_1000_short} and \textbt{efficiency} in Table~\ref{tb:speed_up}.
Fig.~\ref{fig:case_study_short} shows a case where ours can achieve fewer iterative steps when optimizing Eq.~\eqref{eq:obj}.
In Table~\ref{tb:table_1000_short}, ours with BLSOne and TWD achieve comparable low MRE with the SciPy solver \cite{scipy} (supports only CPU and not differentiable) and outperform the vanilla PGD and PGD+RM.
With exploited structures including Prop.~\ref{prop:less_hessian}, ours is faster and more memory efficient than without those structures (``AutoDiff").
More results are in the Appendix.

\begin{table}[!ht]
\vspace{-0.8mm}
\centering
\caption{\textit{With 1,000 random data from $\mathcal{N}(0,1)$.
``In" and ``Out" refer to the number of failed cases inner and outer of the sphere respectively.
``Imp." is the number of cases with FPD $\leq$ the SciPy baseline.
See the Appendix for more results.}
}
\label{tb:table_1000_short}
\setlength{\tabcolsep}{1.5pt}
\resizebox{0.48\textwidth}{!}{
\begin{tabular}{l|rrrr|rrrr}
\hline
\multicolumn{1}{c|}{\multirow{2}{*}{Method}} & \multicolumn{4}{c|}{Size 2$\times$2} & \multicolumn{4}{c}{Size 64$\times$32} \\
 & In$\downarrow$ & Out$\downarrow$ & Imp.$\uparrow$ & MRE$\downarrow$ & In$\downarrow$ & Out$\downarrow$ & Imp.$\uparrow$ & MRE$\downarrow$ \\
\hline
PGD & \underline{172} & 353 & 389 & 10.84 & 548 & \underline{440} & 52 & \underline{0.15} \\
+RM & \textbf{0} & 118 & \underline{844} & 4.30 & 30 & \textbf{0} & \underline{980} & \textbf{0.00} \\
\rowcolor{lightgray} +RM+BLSOne & \textbf{0} & \underline{14} & \textbf{938} & \underline{0.14} & \textbf{6} & \textbf{0} & \textbf{994} & \textbf{0.00} \\
\rowcolor{lightgray} +RM+TWD & \textbf{0} & \textbf{0} & 806 & \textbf{0.00} & \underline{29} & \textbf{0} & 954 & \textbf{0.00} \\
\hline
\end{tabular}
}
\end{table}

\begin{table}[!ht]
\centering
\small
\caption{\textit{Backward speedup with exploited structures.
Numbers are averaged over 100 samples.
``PGD+RM+TWD" is used for LESS.
``Speedup" is (``AutoDiff"-``LESS")/``LESS".}
}
\label{tb:speed_up}
\centering
\setlength{\tabcolsep}{3.5pt}
\resizebox{0.48\textwidth}{!}{
\begin{tabular}{l|cc|cc}
\hline
& \multicolumn{2}{c|}{Size 256 $\times$8$\times$64}
& \multicolumn{2}{c}{Size 256$\times$16$\times$128} \\
& Time (s) & Memory (MB) & Time (s) & Memory (MB) \\
\hline
AutoDiff & 4.44 & 65.22 & 8.64 & 519.32 \\
\rowcolor{lightgray} LESS & 0.01 & 49.20 & 0.02 & 325.70 \\
\hline
\textbf{Speedup} & $\times$444$\uparrow$ & $\times$0.33$\uparrow$ & $\times$432$\uparrow$ & $\times$0.59$\uparrow$ \\
\hline
\end{tabular}
}
\end{table}

\begin{figure}[!ht]
\centering
\begin{subfigure}{0.23\textwidth}
    \includegraphics[width=\textwidth]{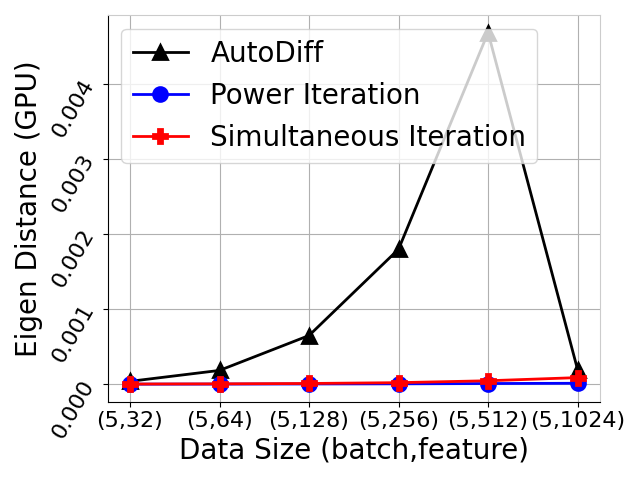}
\end{subfigure}
\hfill
\begin{subfigure}{0.23\textwidth}
    \includegraphics[width=\textwidth]{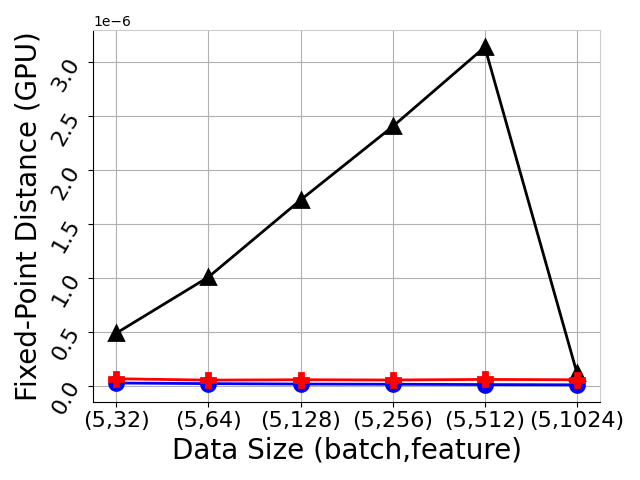}
\end{subfigure}
\vspace{-1mm}
\caption{\textit{Precision evaluation with symmetric $\mb{A}$ in float32.
See the Appendix for nonsymmetric $\mb{A}$ and simulation on ResNet50.}}
\label{fig:ied_symmetric_precision_short}
\vspace{-1mm}
\end{figure}

\textbf{4.2. Evaluation on IED.}
We show the \textbt{effectiveness} in Fig.~\ref{fig:ied_symmetric_precision_short} and \textbt{efficiency} in Table~\ref{tb:ied_symmetric_tease}.
In Fig.~\ref{fig:ied_symmetric_precision_short}, PI and SI achieve lower eigen distance and FPD than PyTorch \texttt{eigh}() (``AutoDiff") \cite{pytorch}.
With exploited structures in Table~\ref{tb:ied_symmetric_tease}, ours achieve higher speed or less memory without ignoring their effectiveness.
More results are in the Appendix.

\begin{table}[!ht]
\centering
\small
\caption{\textit{Efficiency evaluation with symmetric $\mb{A}$ and solution size 5 $\times$ 512.
``J": autodiff Jacobian without exploited structures;
``E": our differentiation with exploited structures.
See the Appendix for results with nonsymmetric $\mb{A}$ and more sizes.}
}
\label{tb:ied_symmetric_tease}
\setlength{\tabcolsep}{6.5pt}
\resizebox{0.49\textwidth}{!}{
\begin{tabular}{l|rr|rr}
\hline
\multicolumn{1}{c|}{\multirow{2}{*}{Method}} & \multicolumn{2}{c|}{\multirow{1}{*}{GPU Time (s)}} & \multicolumn{2}{c}{\multirow{1}{*}{GPU Memory (MB)}} \\
 & Forward & Backward & Forward & Backward \\
\hline
AutoDiff & 0.0758 & 0.0005 & 20.56 & 24.99 \\
\hline
PI-DDN-J & 0.0069 & 0.0436 & 5.55 & 2570.06 \\
PI-IFT-J & 0.0070 & 0.6942 & 5.55 & 5125.00 \\
\rowcolor{lightgray} PI-DDN-E & 0.0069 & 0.0114 & 5.55 & 16.01 \\
\rowcolor{lightgray} PI-IFT-E & 0.0069 & 0.0043 & 5.55 & 31.01 \\
\hline
SI-DDN-J & 1.0623 & 0.0419 & 32.02 & 2570.06 \\
SI-IFT-J & 1.0624 & 0.7120 & 32.02 & 5125.00 \\
\rowcolor{lightgray} SI-DDN-E & 1.0622 & 0.0113 & 32.02 & 16.01 \\
\rowcolor{lightgray} SI-IFT-E & 1.0625 & 0.0042 & 32.02 & 31.01 \\
\hline
\end{tabular}
}
\end{table}

\section{Conclusion}
We explore two deep layers, LESS and IED, for PMaF that can be used in end-to-end learning.
Either optimization improvements or alternatives enhance the optimization effectiveness and/or efficiency.
We also enable the differentiation feasibility and largely reduce computational requirements using
DDN and IFT based gradients with exploited structures.
Extensive experiments show the benefits of our layers over the baseline from SciPy optimizer or PyTorch function.

\clearpage


\bibliography{reference}
\bibliographystyle{icml2023}

\include{appendix}

\end{document}

%% file: appendix.tex
\onecolumn

\section*{Appendix}
\renewcommand{\thesubsection}{\Alph{subsection}}


\subsection{Evaluation Standards}
\textit{\textbf{1) Fixed Point Distance (FPD).}}
This measures the distance of the estimated solution to the minimum of the objective function, where the minimum objective
value is unnecessary to be zero given problem constraints.
\begin{align}
\label{eq:fpd}
\textit{LESS}: \quad &\texttt{F}(\mb{y}) = \norm{ \mb{A} \mb{y} - \mb{b} }\ , \\
\textit{IED}: \quad &\texttt{F}(\mb{y}) = \normx{ \mb{y} - \frac{\mb{A} \mb{y}}{\norm{ \mb{A} \mb{y} }} }\ .
\end{align}

\textit{\textbf{2) Eigen Distance.}}
This is intuitive from Eq.~\eqref{eq:eigen_attribute} to evaluate the eigen solution,
\begin{equation}
\label{eq:edist}
D(\mb{y}) = \norm{ \mb{A} \mb{y} - \lambda \mb{y} }\ .
\end{equation}

\textit{\textbf{3) Mean Relative Error (MRE).}}
Given a reference (generally ground truth) $\mb{y}^r$ and $k^{\text{th}}$ sample, the mean relative error is
\begin{equation}
\label{eq:mre}
\texttt{MRE}(\mb{y},\mb{y}^r) = \frac{1}{K} \sum^K_{k=1} \frac{\texttt{F}(\mb{y}_k) - \texttt{F}(\mb{y}^r_k)}{\texttt{F}(\mb{y}^r_k)} \times 100 \%\ ,
\end{equation}
which can be negative when the estimation $\mb{y}$ outperforms the reference $\mb{y}^r$.

\subsection{Projected Gradient Descent and Riemannian Manifold Projection}
\begin{figure}[!ht]
\centering
\includegraphics[width=0.18\textwidth]{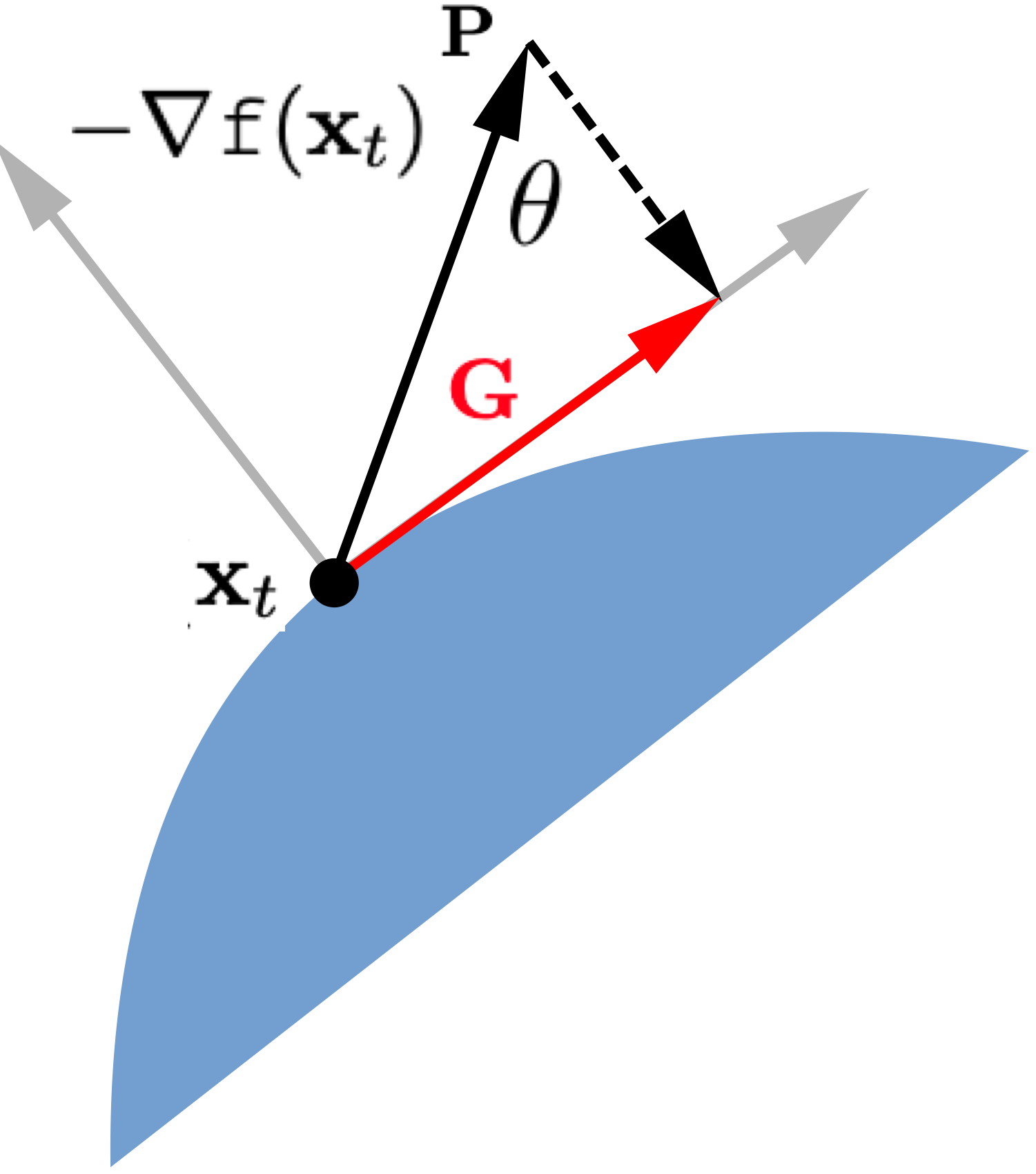}
\caption{\textit{Descent direction projected onto a Riemannian manifold for the sphere constraints.
$-\nabla \texttt{f}(\mb{x}_t)$ is the PGD direction,
$\mb{G}=\text{Proj}_{\texttt{RM}}(-\nabla \texttt{f}(\mb{x}_t))$ is the descent direction on the Riemannian manifold, and $\mb{x}_t$ is on the Riemannian manifold.}}
\label{fig:rie_project}
\end{figure}

\subsubsection{Projected Gradient Descent}
For ease of optimization, the solution is initialized with the least squares solution without constraints as
\begin{align}
\mb{u}_0 &= \left( \mb{A}^\top\mb{A} \right)\inv \mb{A}^\top \mb{b}\ , \label{eq:origin} \\
\mb{u}_{0} &\leftarrow \texttt{Proj}_\texttt{Sph}(\mb{u}_{0})\ .
\end{align}
Then, the solution is iteratively updated in the gradient descent direction $\Delta \mb{u}_t$ at $t^{\text{th}}$ iteration by following
\begin{align}
\mb{u}_{t+1} &= \mb{u}_t +\eta \Delta \mb{u}_t\ , \\
\mb{u}_{t+1} &\leftarrow \texttt{Proj}_{\texttt{Sph}}(\mb{u}_{t+1})
\end{align}
with
\begin{align}
\Delta \mb{u}_t = -\nabla \texttt{f}(\mb{u}_t)\ ,
\quad
\eta = \frac{\norm{\mb{A}^\top (\mb{A}\mb{u}_t-\mb{b})}^2}{\norm{\mb{A}\mb{A}^\top(\mb{A}\mb{u}_t-\mb{b})}^2}\ ,
\end{align}
where $\eta$ is derived by setting $\nabla \texttt{f}(\eta) \equiv \nabla \texttt{f}(\mb{u}_{t+1})=0$ and $\texttt{Proj}_{\texttt{Sph}}(\mb{u})=\mb{u}/\norm{\mb{u}}$ is the projection onto the unit sphere.
The solution update terminates when the objective value decrease is no greater than a predefined tolerance, $1.0e^{-7}$ in our case, or when it reaches the maximum number of iterations, 100 in our case.

\subsubsection{Riemannian Manifold Projection}
Since the solution is constrained on the sphere, a simple projection onto the sphere via gradient descent is inefficient considering the inactive orthogonal descent component.
In Fig.~\ref{fig:rie_project}, considering only the descent component on the tangent plane benefits the solution optimality and optimization process, comparing ``PGD" and ``PGD+RM" in Table~\ref{tb:table_1000}.
\begin{equation}
\begin{aligned}
\label{eq:P}
\mb{P} = \frac{-\mb{x}_t}{\norm{\mb{x}_t}} \norm{\nabla \texttt{f}(\mb{x}_t)} \text{cos}\theta
= \frac{-\mb{x}_t}{\norm{\mb{x}_t}} \norm{\nabla \texttt{f}(\mb{x}_t)} \frac{\nabla \texttt{f}(\mb{x}_t) \frac{-\mb{x}_t}{\norm{\mb{x}_t}}}{\norm{\nabla \texttt{f}(\mb{x}_t)} \normx{ \frac{\mb{x}_t}{\norm{\mb{x}_t}}}}
= \frac{\mb{x}_t \mb{x}^\top_t}{\norm{ \mb{x}_t }^2} \nabla \texttt{f}(\mb{x}_t)\ ,
\end{aligned}
\end{equation}
then
\begin{equation}
\begin{aligned}
\label{eq:G}
\mb{G}
= -\nabla \texttt{f}(\mb{x}_t) + \mb{P}
= -\nabla \texttt{f}(\mb{x}_t) + \frac{\mb{x}_t \mb{x}^\top_t}{\norm{ \mb{x}_t }^2} \nabla \texttt{f}(\mb{x}_t)\ .
\end{aligned}
\end{equation}

Hence,
\begin{equation}
\begin{aligned}
\label{eq:rie_detail}
\texttt{Proj}_{\texttt{RM}}\left( -\nabla \texttt{f}(\mb{x}_t) \right)
= \left( \mb{I}_n - \frac{\mb{x}_t \mb{x}^\top_t}{\norm{ \mb{x}_t }^2} \right) \left( -\nabla \texttt{f}(\mb{x}_t) \right)\ ,
\end{aligned}
\end{equation}
where $\mb{x}_t \in \mathbb{R}^n$, $\nabla \texttt{f}(\mb{x}_t) \in \mathbb{R}^n$, $\mb{I}_n$ is an $n \times n$ identity matrix.

\subsection{Gradient Transformation with Flipped Eigenvalues}
For different orderings of multiple eigenvalues, for instance, descending-order (concerning the order of eigenvalues) versus ascending-order eigenvalues, the gradients of their associated eigenvectors need to be flipped according to the flipping of the eigenvalues.
Recall the eigen decomposition problem defined in Eq.~\eqref{eq:ed}.
Reserve all of the eigenvalues of $\mb{A}$, that is $n=m$, a flipping function is defined as $\texttt{g}(\mb{x}): \mathbb{R}^{m \times m} \rightarrow \mathbb{R}^{m \times m}$, that is
\begin{equation}
\label{eq:flip}
\texttt{g}(\mb{x}) =
\begin{bmatrix}
 0 & \cdots &  1 \\
 \vdots & \iddots & \vdots \\
1 & \cdots & 0
\end{bmatrix}
\mb{x}
\begin{bmatrix}
 0 & \cdots &  1 \\
 \vdots & \iddots & \vdots \\
1 & \cdots & 0
\end{bmatrix} \\
=
\mb{I}^{\ast}_{m} \mb{x} \mb{I}^{\ast}_{m}\ ,
\end{equation}
where $(\mb{I}^{\ast}_n \mb{x})$ flips $\mb{x}$ in the vertical direction and $(\mb{x} \mb{I}^{\ast}_n)$ in the horizontal direction.
For the notation simplicity, we denote $\texttt{g}(\mb{x})=\mb{x}^{\text{flip}}$.
Then,
\begin{align}
    \mb{A} \mb{u}
    &= \mb{\Lambda} \mb{u}
    = \mb{I}^{\text{flip}}_m \mb{\Lambda}^{\text{flip}} \mb{I}^{\text{flip}}_m \mb{u}\ , \\
    \mb{I}^{\text{flip}}_m \mb{A} \mb{u}
    &= \mb{\Lambda}^{\text{flip}} \mb{I}^{\text{flip}}_m \mb{u}\ , \\
    \mb{I}^{\text{flip}}_m \mb{A} \left( \mb{I}^{\text{flip}}_m \mb{u}^{\text{flip}} \mb{I}^{\text{flip}}_m \right)
    &= \mb{\Lambda}^{\text{flip}} \mb{u}^{\text{flip}} \mb{I}^{\text{flip}}_m\ , \\
    \mb{A}^{\text{flip}} \mb{u}^{\text{flip}}
    &= \mb{\Lambda}^{\text{flip}} \mb{u}^{\text{flip}}\ ,
\end{align}
where the input $\mb{A} \in \mathbb{R}^{m \times m}$, eigenvectors $\mb{u} \in \mathbb{R}^{m \times m}$, and eigenvalue matrix $\mb{\Lambda} \in \mathbb{R}^{m \times m}$.
It indicates that to obtain the reverse-order eigenvalues, $\mb{A}$ needs to be flipped, which also leads to the flipping of
$\mb{u}$.
Therefore, to compare the descending-order eigen decomposition with an ascending-order one, both $\mb{A}$ and $\mb{u}$ need to be flipped.

For the gradient of loss $L$ over the flipped eigenvectors $\mb{u}^{\text{flip}}$, that is $\nabla_{\mb{u}^{\text{flip}}} L$, since $\mb{u}^{\text{flip}}=\mb{I}^{\text{flip}}_n \mb{u} \mb{I}^{\text{flip}}_n$,
\begin{equation}
    \nabla_{\mb{u}^{\text{flip}}} L
    = \mb{I}^{\text{flip}}_n \nabla_{\mb{u}^{\text{flip}}} L \mb{I}^{\text{flip}}_n
    = \left( \nabla_{\mb{u}} L \right)^{\text{flip}} \ .
\end{equation}

In sum, for the reverse-order eigenvalues, gradients of their associated eigenvectors need to be flipped in both the horizontal and vertical directions, that is to apply Eq.~\eqref{eq:flip}, when the eigenvalue order (or equivalently the eigenvector order) is flipped.

\subsection{Proof and Demo Code for Proposition~\ref{prop:less_hessian}}
We explore the matrix structure by taking an example with $m=2$ and $n=3$, that is $\mb{A}$, $\mb{b}$, and $\mb{y}$ as
\begin{equation}
\mb{A} =
\begin{bmatrix}
a_{00} & a_{01} & a_{02} \\
a_{10} & a_{11} & a_{12}
\end{bmatrix}\ ,
\quad
\mb{b} =
\begin{bmatrix}
b_0 \\
b_1
\end{bmatrix}\ , \\
\quad
\mb{y} =
\begin{bmatrix}
y_0 \\
y_1 \\
y_2
\end{bmatrix}\ .
\end{equation}
The Jacobian vector and Hessian matrices of the objective function in Eq.~\eqref{eq:obj} over the solution $\mb{y}$ and the Hessian matrix over entries $\mb{A}$ and $\mb{b}$ are
\begin{align}
\nabla_Y \texttt{f}(\mb{A}, \mb{b}, \mb{y})
&= \mb{A}^\top \left( \mb{A} \mb{y} - \mb{b} \right) \nonumber \\
&=
\begin{bmatrix}
(a_{00}^2 + a_{10}^2) y_0 + (a_{00} a_{01} + a_{10} a_{11}) y_1 + (a_{00} a_{02} + a_{10} a_{12}) y_2 \\
(a_{01} a_{00} + a_{11} a_{10}) y_0 + (a_{01}^2 + a_{11}^2) y_1 + (a_{01} a_{02} + a_{11} a_{12}) y_2 \\
(a_{02} a_{00} + a_{12} a_{10}) y_0 + (a_{02} a_{01} + a_{12} a_{11} y_1 + (a_{02}^2 + a_{12}^2) y_2)
\end{bmatrix}
-
\begin{bmatrix}
a_{00} b_0+ a_{10} b_1 \\
a_{01} b_0 + a_{11} b_1 \\
a_{02} b_0 + a_{12} b_1
\end{bmatrix}\ ,\\
\nabla^2_{bY} \texttt{f}(\mb{A}, \mb{b}, \mb{y})
&=
-\begin{bmatrix}
a_{00} & a_{10} \\
a_{01} & a_{11} \\
a_{02} & a_{12}
\end{bmatrix}
=-\mb{A}^\top\ , \\
\nabla^2_{AY} \texttt{f}(\mb{A}, \mb{b}, \mb{y})
&=
\begin{bmatrix}
\begin{bmatrix}
2a_{00} y_0 + a_{01} y_1 + a_{02} y_2 & a_{00} y_1 & a_{00} y_2 \\
2a_{10} y_0 + a_{11} y_1 + a_{12} y_2 & a_{10} y_1 & a_{10} y_2
\end{bmatrix} \\ \\
\begin{bmatrix}
a_{01} y_0 & a_{00} y_0 + 2a_{01} y_1 + a_{02} x_2 & a_{01} y_2 \\
a_{11} y_0 & a_{10} y_0 + 2a_{11} y_1 + a_{12} x_2 &  a_{11} y_2
\end{bmatrix} \\ \\
\begin{bmatrix}
a_{02} y_0 & a_{02} y_1 & a_{00} y_0 + a_{01} y_1 + 2a_{02} y_2 \\
a_{12} y_0 & a_{12} y_1 & a_{10} y_0 + a_{11} y_1 + 2a_{12} y_2
\end{bmatrix}
\end{bmatrix}
-
\begin{bmatrix}
\begin{bmatrix}
b_0 & 0 & 0 \\
b_1 & 0 & 0
\end{bmatrix} \\ \\
\begin{bmatrix}
0 & b_0 & 0\\
0 & b_1 & 0
\end{bmatrix} \\ \\
\begin{bmatrix}
0 & 0 & b_0\\
0 & 0 & b_1
\end{bmatrix}
\end{bmatrix} \nonumber \\
&=
\begin{bmatrix}
\begin{bmatrix}
a_{00} y_0 & a_{00} y_1 & a_{00} y_2 \\
a_{10} y_0 & a_{10} y_1 & a_{10} y_2
\end{bmatrix}\\ \\
\begin{bmatrix}
a_{01} y_0 & a_{01} y_1 & a_{01} y_2 \\
a_{11} y_0 & a_{11} y_1 & a_{11} y_2
\end{bmatrix}\\ \\
\begin{bmatrix}
a_{02} y_0 & a_{02} y_1 & a_{02} y_2 \\
a_{12} y_0 & a_{12} y_1 & a_{12} y_2
\end{bmatrix}
\end{bmatrix}
+
\begin{bmatrix}
\begin{bmatrix}
a_{00} y_0 + a_{01} y_1 + a_{02} y_2 & 0 & 0 \\
a_{10} y_0 + a_{11} y_1 + a_{12} y_2 & 0 & 0
\end{bmatrix}\\ \\
\begin{bmatrix}
0 & a_{00} y_0 + a_{01} y_1 + a_{02} y_2 & 0 \\
0 & a_{10} y_0 + a_{11} y_1 + a_{12} y_2 & 0
\end{bmatrix}\\ \\
\begin{bmatrix}
0 & 0 & a_{00} y_0 + a_{01} y_1 + a_{02} y_2 \\
0 & 0 & a_{10} y_0 + a_{11} y_1 + a_{12} y_2
\end{bmatrix}
\end{bmatrix}
-
\begin{bmatrix}
\begin{bmatrix}
b_0 & 0 & 0 \\
b_1 & 0 & 0
\end{bmatrix} \\ \\
\begin{bmatrix}
0 & b_0 & 0\\
0 & b_1 & 0
\end{bmatrix} \\ \\
\begin{bmatrix}
0 & 0 & b_0\\
0 & 0 & b_1
\end{bmatrix}
\end{bmatrix} \nonumber \\
&=
\begin{bmatrix}
\begin{bmatrix}
a_{00} \\
a_{10}
\end{bmatrix} \mb{y}^\top \\ \\
\begin{bmatrix}
a_{01} \\
a_{11}
\end{bmatrix} \mb{y}^\top \\ \\
\begin{bmatrix}
a_{02} \\
a_{12}
\end{bmatrix} \mb{y}^\top
\end{bmatrix}
+
\begin{bmatrix}
\mb{A} \mb{y} & \mb{0} & \mb{0} \\
\mb{0} & \mb{A} \mb{y} & \mb{0} \\
\mb{0} & \mb{0} & \mb{A} \mb{y}
\end{bmatrix}
-
\begin{bmatrix}
\mb{b} & \mb{0} & \mb{0} \\
\mb{0} & \mb{b} & \mb{0} \\
\mb{0} & \mb{0} & \mb{b}
\end{bmatrix} \nonumber \\
&=
\begin{bmatrix}
\mb{A}_0 \mb{y}^\top \\
\mb{A}_1 \mb{y}^\top \\
\mb{A}_2 \mb{y}^\top
\end{bmatrix}
+
\begin{bmatrix}
\mb{A} \mb{y} - \mb{b} & \mb{0} & \mb{0} \\
\mb{0} & \mb{A} \mb{y} - \mb{b} & \mb{0} \\
\mb{0} & \mb{0} & \mb{A} \mb{y} - \mb{b}
\end{bmatrix}\ ,
\end{align}
where the left term can be written as $\mb{A}_i$ for all $i \in \mc{N} = \{1, ..., n\}$ and
the right term assigns $(\mb{A} \mb{y} - \mb{b})$ to the diagonal of $\nabla^2_{AY} \texttt{f}(\mb{A}, \mb{b}, \mb{y})$ which is $\mc{B}$ in Eq.~\eqref{eq:hessian_less}.
One can represent it as
\begin{align}
    \mc{B}_{ij} &= \mb{A}_i \mb{y}^\top_j, \quad \forall i, j \in \mc{N}\ ,\\
    \mc{B}_{ii} &\leftarrow \mc{B}_{ii} + (\mb{A} \mb{y} - \mb{b}), \quad \forall i \in \mc{N}\ .
\end{align}
One can also obtain the same structure from different examples.
It is clear that forming $\mc{B}$ using $\mb{A}$, $\mb{b}$, and $\mb{y}$ in this way avoids calculating the zero vectors and matrices, and thus, becoming more efficient than using Jacobian to calculate every element of such a spare matrix.
To this end, we arrive at the result in Prop.~\ref{prop:less_hessian}. $\square$

\clearpage
Meanwhile, we show an implementation example comparing the automatic Jacobian (without exploited structures) and ours with the exploited structure in Listing~\ref{lst:second_order}.
\begin{lstlisting}[basicstyle=\footnotesize\ttfamily,linewidth=16cm,label={lst:second_order},caption={\textit{\normalsize \hspace{8mm}An implementation example of exploited Hessian structure for LESS.}\\}]
import torch
from torch.autograd import grad
from torch.autograd.functional import jacobian as J

def obj_fnc(A, u, b):
    Au_b = torch.einsum('bmn,bn->bm', A, u) - b
    loss = 0.5 * torch.einsum('bm,bm->b', Au_b, Au_b)

    return loss.sum()

# ==== Autogradient
def dfdy_auto_fnc(A, u, b):
    return grad(obj_fnc(A, u, b), (u), create_graph=True)[0]

def dffdAy_auto_fnc(A, u, b):
    m, n = A.shape[1:3]
    D = []

    with torch.enable_grad():
        for A_p, u_p, b_p in zip(A, u, b):
            A_p = A_p.view(1, m, n)
            u_p = u_p.view(1, n)
            b_p = b_p.view(1, m)
            D.append(J(lambda A: dfdy_auto_fnc(A, u_p, b_p), (A_p)))

        D = torch.cat(D, dim=0)

    return D

# ==== Structure exploited
def dfdy_fnc(A, u, b):
    Au_b = torch.einsum('bmn,bn->bm', A, u) - b

    return torch.einsum('bmn,bm->bn', A, Au_b)

def dffdAy_fnc(A, u, b):
    batch, m, n = A.shape
    D = A.new_zeros(batch, n, m, n)

    with torch.no_grad():
        for i in range(n):
            D_1 = torch.einsum('bm,bn->bmn', A[:, :, i], u)
            D_2 = torch.einsum('bmn,bn->bm', A, u) - b
            D[:, i] = D_1
            D[:, i, :, i] += D_2

    return D
\end{lstlisting}

\subsection{Proof for Proposition~\ref{prop:ied_gradient}}
For the principal matrix features, the implicit differentiation is on the eigenvector associated with the largest eigenvalue as
\begin{equation}
\nabla_X L(\mathbf{y})
= \nabla_Y L(\mathbf{y}) \nabla_X \mathbf{y}
= \nabla_Y L(\mathbf{y}) 
\left( \mc{H}\inv \mc{A}^{T} \left( \mc{A} \mc{H}\inv \mc{A}^\top \right)\inv \mc{A} \mc{H}\inv - \mc{H}\inv \right) \mc{B}
= \mc{K} \mc{B} \in \mathbb{R}^{m \times m}\ ,
\end{equation}
where $\mc{K} \in \mathbb{R}^{m}$ is a vector given the upper-level loss $L \in \mathbb{R}$ from the bi-level problem.
Then,
\begin{equation}
\label{eq:prop_34_app}
\nabla_X L(\mathbf{y}) = -\mc{K} \mathbf{y}^\top - \mathbf{y} \mc{K}^\top\ .
\end{equation}

We find the matrix structure by taking the example of $m=2$, where $\mc{K}=[ k_0, k_1 ]^\top$ and $\mb{y} = [ y_0, y_1 ]^\top$.
Then,
\begin{equation}
\mc{B}
= \mc{B}_0 + \mc{B}_1
= \begin{bmatrix}
-y_0 & -y_1 & 0 & 0 \\
0 & 0 & -y_0 & -y_1
\end{bmatrix}
+
\begin{bmatrix}
-y_0 & 0 & -y_1 & 0 \\
0 & -y_0 & 0 & -y_1
\end{bmatrix}
\end{equation}
and
\begin{equation}
\mc{K} \mc{B}
= \mc{K} \mc{B}_0 + \mc{K} \mc{B}_1 =
\begin{bmatrix}
-k_0 y_0 & -k_0 y_1 \\
-k_1 y_0 & -k_1 y_1
\end{bmatrix}
+
\begin{bmatrix}
-k_0 y_0 & -k_1 y_0 \\
-k_0 y_1 & -k_1 y_1
\end{bmatrix}
=
-\begin{bmatrix}
k_0\\
k_1
\end{bmatrix}
\begin{bmatrix}
y_0 & y_1
\end{bmatrix}
-
\begin{bmatrix}
y_0\\
y_1
\end{bmatrix}
\begin{bmatrix}
k_0 & k_1
\end{bmatrix}
=
- \mc{K} \mb{y}^\top - \mb{y} \mc{K}^\top\ ,
\end{equation}
where $\mc{K} \mc{B}_0$ is the outer-product of $\mc{K}$ and $(-\mb{y})$, both $\mc{K} \mc{B}_1$ and $\mc{K} \mc{B}_0$ will be reshaped to $m \times m$ matrices.
One can also apply the same rule to different examples, which will have the same
structure in Eq.~\eqref{eq:prop_34_app}.
Clearly, using $\mb{y} \in \mathbb{R}^{m}$ for $\nabla_X L(\mathbf{y})$ is much more efficient and requires much less memory than using $\mc{B} \in \mathbb{R}^{m \times (m \times m)}$.
To this end, we prove Prop.~\ref{prop:ied_gradient}. $\square$

\subsection{More Results of LESS Optimization}
\begin{figure*}[!ht]
\centering
\includegraphics[width=\textwidth]{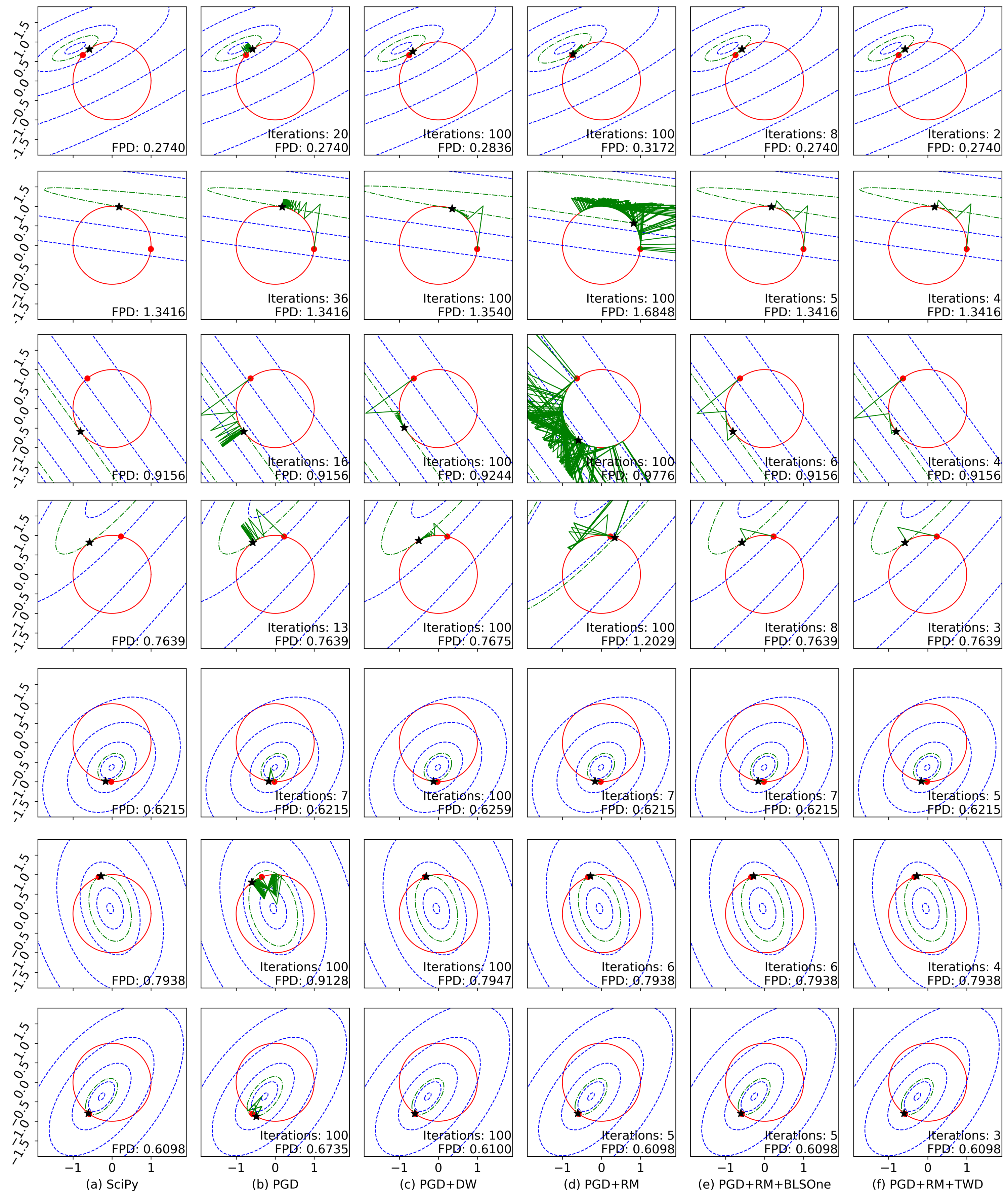}
\vspace{-5mm}
\caption{\textit{Illustration of the iterative optimization in LESS.
Each row is a sample with 6 methods.
``{\color{blue}\textbf{blue}} dash": the least squares function;
``{\color{red}\textbf{red}} solid": the sphere constraints;
``{\color{green}\textbf{green}} solid": solution updates before and after $\texttt{Proj}_{\texttt{RM}}()$;
``{\color{green}\textbf{green}} dot-dash": the least squares function with the final solution;
``{\color{red}\textbf{red}} dot": the initial solution;
``{\color{black}\textbf{black}} star": the final solution, not always the optimal.
The least squares function with the exact solution ({\color{green}\textbf{green}} dot-dash) should be tangent to the sphere ({\color{red}\textbf{red}} solid).
Our ``PGD+RM+BLSOne" and ``PGD+RM+TWD" require much fewer iterations than the others for the optimal solution with comparable fixed point distance (FPD).}}
\label{fig:case_study_full}
\end{figure*}
In addition to Fig.~\ref{fig:case_study_short}, Fig.~\ref{fig:case_study_full} contains the full list of the compared methods.
The inputs for Fig.~\ref{fig:case_study_short} are
\begin{equation}
\mb{A} =
\begin{bmatrix}
0.569525 & -1.254572 \\
0.414020 & 0.124439
\end{bmatrix}\quad \text{and}
\quad
\mb{b} =
\begin{bmatrix}
-1.583332\\
-0.286124
\end{bmatrix}\ .
\end{equation}
In this case, the vanilla PGD requires more iterations than our ``PGD+RM+BLSOne" and ``PGD+RM+TWD".
Statistically, from Table~\ref{tb:table_1000}, when comparing ``PGD" with ``PGD+RM", all the metrics are better than ``PGD".
With ``+BLSOne" and ``+TWD", these can be further improved.

\begin{table*}[!ht]
\centering
\caption{\textit{Effectiveness evaluation on 1,000 random data sampled from $\mathcal{N}(0, 1)$ normal distribution.
\textbf{Bold}: the best, \underline{underline}: the second best.
``PGD": projected gradient descent (maximum 100 iterations),
``RM": Riemannian manifold,
``BLS": backtracking line search ($\alpha$=0.5 and $\beta$=0.8),
``TWD": tangent weight decay on Riemannian manifold ($\beta$=0.9),
``DW": direction weight,
``In" and ``Out": the number of failed cases inner and outer of the sphere respectively, where the case is regarded as failed when the solution update reaches the maximum 100 iterations,
``Imp.": the number of cases with energy no greater than SciPy.
Problem sizes, m$\times$n for A, are 2$\times$2 (282 inner and 718 outer), 64$\times$32 (555 inner and 445 outer), and 1024$\times$256 (all inner).}
}
\vspace{1mm}
\label{tb:table_1000}
\centering
\setlength{\tabcolsep}{3pt}
\resizebox{\textwidth}{!}{
\begin{tabular}{ccccccc|rrrr|rrrr|rrrr}
\hline
\multirow{2}{*}{SciPy} & \multirow{2}{*}{PGD} & \multirow{2}{*}{RM} & \multicolumn{2}{c}{BLS} & \multirow{2}{*}{TWD} & \multirow{2}{*}{DW} & \multicolumn{4}{c|}{2$\times$2} & \multicolumn{4}{c|}{64$\times$32} & \multicolumn{4}{c}{1024$\times$256} \\
& & & $\eta=1$ & $\eta=\nabla$ & & & In$\downarrow$ & Out$\downarrow$ & Imp.$\uparrow$ & MRE$\downarrow$ & In$\downarrow$ & Out$\downarrow$ & Imp.$\uparrow$ & MRE$\downarrow$ & In$\downarrow$ & Out$\downarrow$ & Imp.$\uparrow$ & MRE$\downarrow$ \\
\hline
\checkmark & & & & & & & \multicolumn{1}{c}{-} & \multicolumn{1}{c}{-} & \multicolumn{1}{c}{-} & \multicolumn{1}{c|}{-} & \multicolumn{1}{c}{-} & \multicolumn{1}{c}{-} & \multicolumn{1}{c}{-} & \multicolumn{1}{c|}{-} & \multicolumn{1}{c}{-} & \multicolumn{1}{c}{-} & \multicolumn{1}{c}{-} & \multicolumn{1}{c}{-} \\

& \checkmark & & & & & & \underline{172} & 353 & 389 & 10.84 & 548 & 440 & 52 & 0.15 & 1000 & \multicolumn{1}{c}{-} & 0 & 1.35 \\

& \checkmark & & & & & \checkmark & 179 & 438 & 37 & 4.89 & 474 & 363 & 63 & 0.07 & 1000 & \multicolumn{1}{c}{-} & 0 & 0.47 \\

& \checkmark & \checkmark & & & & & \textbf{0} & 118 & 844 & 4.30 & 30 & \textbf{0} & \underline{980} & \textbf{0.00} & \underline{4} & \multicolumn{1}{c}{-} & \textbf{1000} & \textbf{-0.03} \\

\rowcolor{lightgray} & \checkmark & \checkmark & \checkmark & & & & \textbf{0} & \underline{14} & \textbf{938} & \underline{0.14} & \textbf{6} & \textbf{0} & \textbf{994} & \textbf{0.00} & \underline{4} & \multicolumn{1}{c}{-} & \textbf{1000} & \textbf{-0.03} \\

& \checkmark & \checkmark & \checkmark & & & \checkmark & 175 & 421 & 40 & 3.51 & 285 & \underline{142} & 286 & \underline{0.01} & 984 & \multicolumn{1}{c}{-} & \underline{999} & \textbf{-0.03} \\

& \checkmark & \checkmark & & \checkmark & & & \textbf{0} & 108 & \underline{853} & 4.21 & \underline{7} & \textbf{0} & \textbf{994} & \textbf{0.00} & \textbf{2} & \multicolumn{1}{c}{-} & \textbf{1000} & \textbf{-0.03} \\

& \checkmark & \checkmark & & \checkmark & & \checkmark & 204 & 536 & 17 & 1.19 & 472 & 355 & 92 & 0.04 & 1000 & \multicolumn{1}{c}{-} & 0 & \underline{0.42} \\

\rowcolor{lightgray} & \checkmark & \checkmark & & & \checkmark & & \textbf{0} & \textbf{0} & 806 & \textbf{0.00} & 29 & \textbf{0} & 954 & \textbf{0.00} & \underline{4} & \multicolumn{1}{c}{-} & \textbf{1000} & \textbf{-0.03} \\

& \checkmark & \checkmark & & & \checkmark & \checkmark & 204 & 533 & 15 & 3.12 & 466 & 352 & 90 & 0.04 & 1000 & \multicolumn{1}{c}{-} & 0 & \underline{0.42} \\
\hline
\end{tabular}
}
\end{table*}

\clearpage
\subsection{Evaluation of IED Precision and Computational Requirements}

\subsubsection{Symmetric and Nonsymmetric Data Sampling}
\label{sec:sampling}
\begin{figure}[!ht]
\centering
\begin{subfigure}{0.24\textwidth}
    \includegraphics[width=\textwidth]{figures/ied_data/solve_symmetric_stopFalse_float32_eigen_gap_gpu.png}
    \caption{ED, float32}
\end{subfigure}
\begin{subfigure}{0.24\textwidth}
    \includegraphics[width=\textwidth]{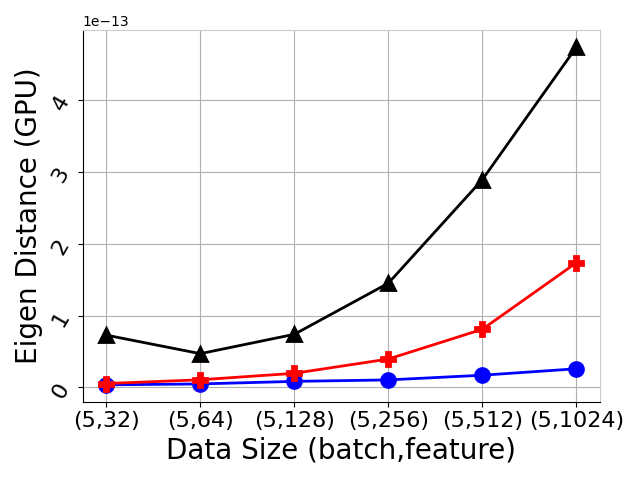}
    \caption{ED, float64}
\end{subfigure}
\begin{subfigure}{0.24\textwidth}
    \includegraphics[width=\textwidth]{figures/ied_data/solve_symmetric_stopFalse_float32_fp_gap_gpu.png}
    \caption{FPD, float32}
\end{subfigure}
\begin{subfigure}{0.24\textwidth}
    \includegraphics[width=\textwidth]{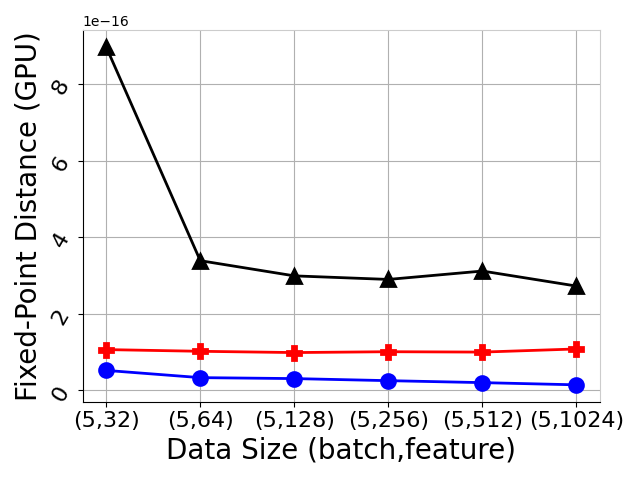}
    \caption{FPD, float64}
\end{subfigure}
\vspace{-3mm}
\caption{\textit{Symmetric $\mb{A}$ sampled from Gaussian distribution with activation, precision evaluation with eigen distance and FPD.}}
\label{fig:ied_symmetric_precision}
\end{figure}

\begin{figure}[!ht]
\centering
\begin{subfigure}{0.24\textwidth}
    \includegraphics[width=\textwidth]{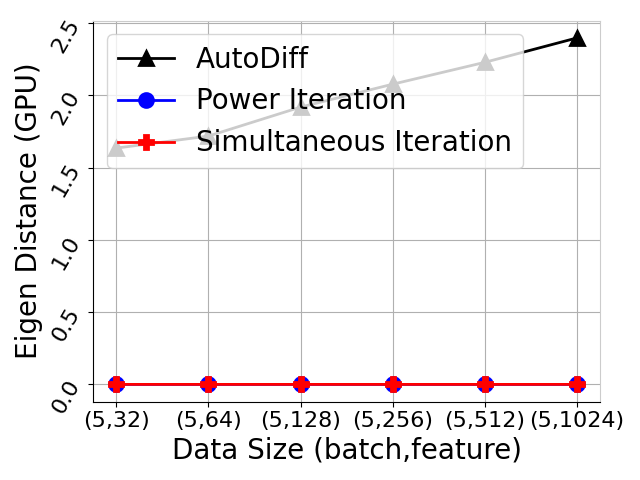}
    \caption{ED, float32}
\end{subfigure}
\begin{subfigure}{0.24\textwidth}
    \includegraphics[width=\textwidth]{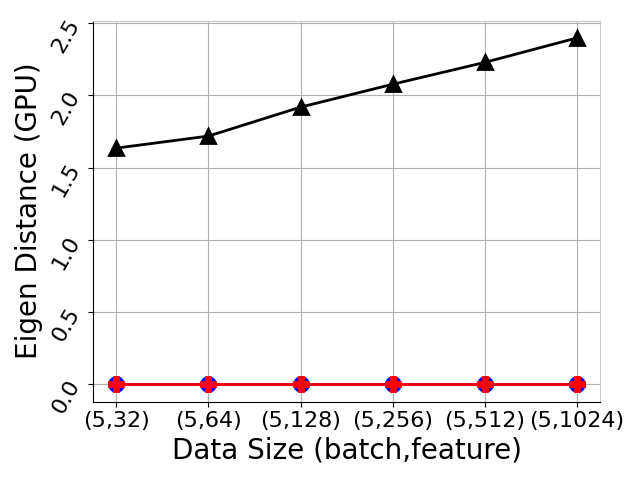}
    \caption{ED, float64}
\end{subfigure}
\begin{subfigure}{0.24\textwidth}
    \includegraphics[width=\textwidth]{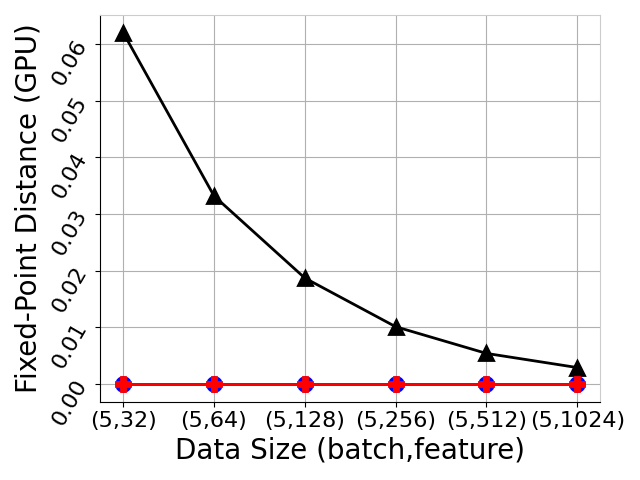}
    \caption{FPD, float32}
\end{subfigure}
\begin{subfigure}{0.24\textwidth}
    \includegraphics[width=\textwidth]{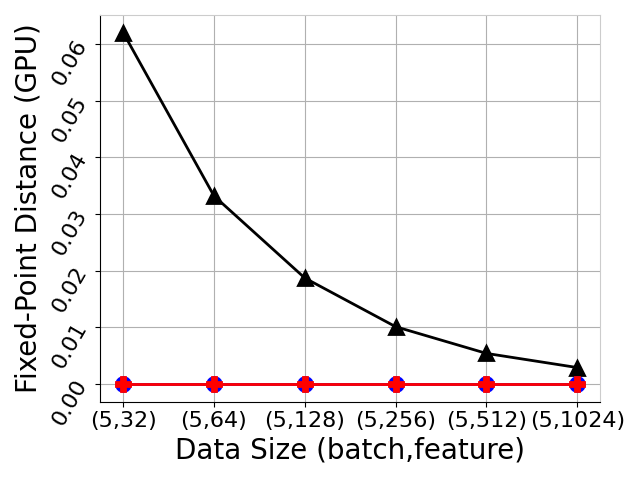}
    \caption{FPD, float64}
\end{subfigure}
\vspace{-3mm}
\caption{\textit{Nonsymmetric $\mb{A}$ sampled from Gaussian distribution with activation, precision evaluation with eigen distance and FPD.}}
\label{fig:ied_nonsymmetric_precision}
\end{figure}

\textbf{Sampling Distributions}. Our evaluated data $\mb{A}$ are sampled from \textit{the standard Gaussian distribution} $\mc{N}(0, 1)$, \textit{uniform distribution} in $[0, 1)$, \textit{von Mises distribution} in $\mc{V}(0, 1)$ \cite{vonmise}, and \textit{random choices} from [0.0, 10.0), with absolute values, where $\mb{A} \leftarrow \mb{A}+\mb{A}^\top$ is applied for symmetry.
We use $\texttt{numpy.random.randn}()$, $\texttt{numpy.random.uniform}()$, $\texttt{numpy.random.vonmises}(0, 1)$, and $\texttt{numpy.random.choice}(10.0)$, respectively, from the NumPy library \cite{harris2020array}.
They achieve similar numerical ranges as that from the Gaussian distribution in Figs.~\ref{fig:ied_symmetric_precision}-\ref{fig:ied_nonsymmetric_precision}.
Hyperparameters required by the sampling methods are not limited to the settings in this work, such as the mean and variance in the von Mises distribution and the value upper bound in the random choices.

\begin{figure}[!ht]
\centering
\begin{subfigure}{0.24\textwidth}
    \includegraphics[width=\textwidth]{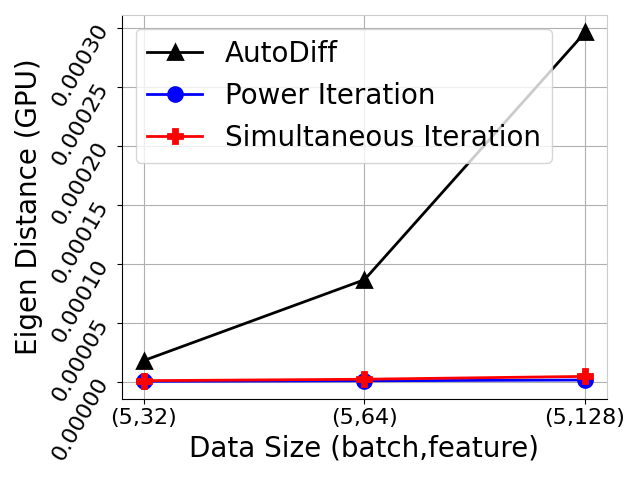}
    \caption{Symmetric $\mb{A}$, ED}
\end{subfigure}
\begin{subfigure}{0.24\textwidth}
    \includegraphics[width=\textwidth]{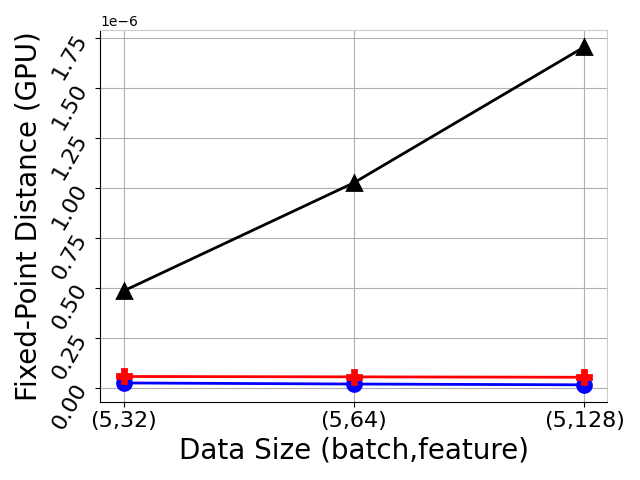}
    \caption{Symmetric $\mb{A}$, FPD}
\end{subfigure}
\begin{subfigure}{0.24\textwidth}
    \includegraphics[width=\textwidth]{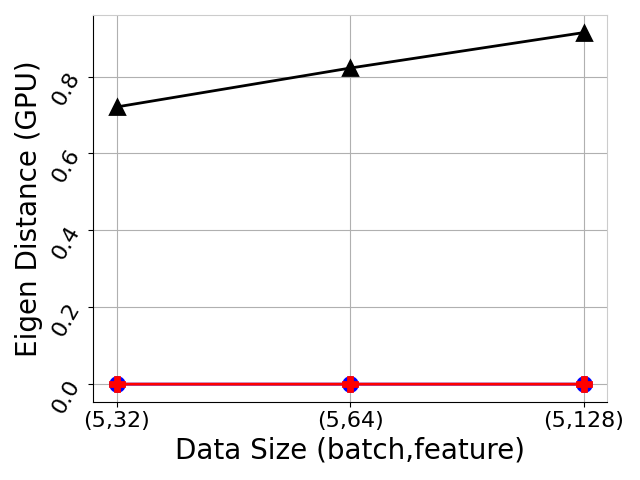}
    \caption{Nonsymmetric $\mb{A}$, ED}
\end{subfigure}
\begin{subfigure}{0.24\textwidth}
    \includegraphics[width=\textwidth]{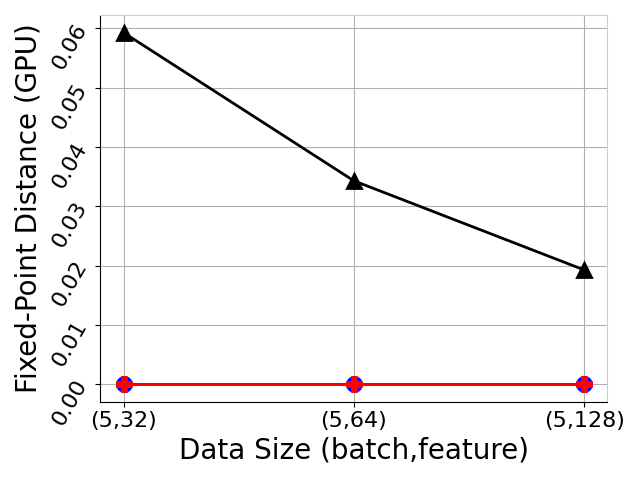}
    \caption{Nonsymmetric $\mb{A}$, FPD}
\end{subfigure}
\vspace{-3mm}
\caption{\textit{Evaluation on 3 data sizes within the memory capacity as it requires a $(2048 \times 8^2) \times m^2$ dimension fully-connected layer in ResNet50, where $m=\{32, 64, 128\}$. All data is in float32.
Both metrics are the less the better.}}
\label{fig:ied_precision_resnet}
\end{figure}

The feasibility lies in 1) the simulation of normalized outputs (such as logits or softmax probability) of neural networks with an absolute operation,
2) the requirement of positive largest eigenvalue for IFT in Eq.~\eqref{eq:ift_pi}.
For nonsymmetric sampling, as we test 10,000 random samples (each with batch size 5) under this setting,
the power iteration and simultaneous iteration algorithms can always achieve almost 0 eigen distance and 0 fixed-point distance.
In contrast, PyTorch \texttt{eigh}()\footnote{PyTorch \texttt{eigh}() was designed for symmetric $\mb{A}$ and PyTorch \texttt{eig}() for complex and general data for nonsymmetric $\mb{A}$, but we are 1) not aim at complex data and 2) to avoid using \texttt{eigh}() for symmetric matrix but rather PI or SI by investigating these results.}, denoted as ``AutoDiff", has a much larger eigen distance, particularly in float32.

\textbf{Simulation on ResNet50}. We further generate random samples by using \textit{a modified ResNet50} with pretrained weights, removing the adaptive average pooling layer and adjusting the last fully-connected layer to obtain $m^2$ out channels (instead of 1,000) which will then be reshaped to a $m \times m$ matrix, followed by the absolute operation and symmetry for $\mc{A}$ if required.
It follows: random data with size ($5 \times 3 \times 256 \times 256$) that is (batch $\times$ channel $\times$ height $\times$ width) sampled from $\mc{N}(0, 1)$ $\rightarrow$ \texttt{ResNet}() $\rightarrow$ reshape to $m \times m$ $\rightarrow$ absolute operation $\rightarrow$ ($\mb{A} \leftarrow \mb{A} + \mb{A}^\top$) if symmetric $\mb{A}$ is required.
In Figs.~\ref{fig:ied_precision_resnet}, ``PI" and ``SI" still achieve better eigen solutions than ``AutoDiff" on both symmetric and nonsymmetric $\mb{A}$.

\subsubsection{Additional Experiments on Computational Requirements}
We evaluate the solution precision with eigen distance and fixed-point distance on symmetric and nonsymmetric $\mb{A}$ in Fig.~\ref{fig:ied_symmetric_precision} and Fig.~\ref{fig:ied_nonsymmetric_precision} respectively.
Additionally, the computational requirements include running time and GPU memory in Tables~\ref{tb:ied_symmetric_time_cpu}-\ref{tb:ied_nonsymmetric_memory} for both symmetric and nonsymmetric $\mb{A}$.
Although ``AutoDiff" achieves faster speed in most cases, its precision is inferior to ``PI" and ``SI", particularly in float32.
For the evaluation completion, we provide all of these results.

Meanwhile, since a solver achieves the same solutions given the same inputs, it consumes the same (at least quite similar) computational resources in the backward pass.
We modulate each solver with different backward propagation methods into individual layers and provide all the forward and backward results.

In all these tables, implicit differentiation without exploited structure (``J") and with our exploited structure (``E") clearly distinguishes the benefits of exploiting the Jacobian and Hessian matrices in both the running time and memory requirements.
For those with out-of-memory issues or extensive running time, we mark the results with ``-" in the tables.

\begin{table*}[!ht]
\centering
\small
\caption{\textit{Implicit eigen decomposition (IED) layer, CPU time (s) for symmetric $\mb{A}$.
``AutoDiff": PyTorch \texttt{eigh}() function;
``DDN": deep declarative network;
``IFT": implicit function theorem;
``unroll": unrolling the forward iteration for gradients via PyTorch autodiff mechanism;
``J": autodiff Jacobian without exploited structure;
``E": our implicit differentiation with exploited structure.
Our \colorbox{gray!50}{best suggestions} are highlighted considering the overall solution precision and computational requirements.}
}
\label{tb:ied_symmetric_time_cpu}
\setlength{\tabcolsep}{11pt}
\resizebox{\textwidth}{!}{
\begin{tabular}{l|rr|rr|rr|rr}
\hline
\multicolumn{1}{c|}{\multirow{2}{*}{Method}} & \multicolumn{2}{c|}{5$\times$32} & \multicolumn{2}{c|}{5$\times$64} & \multicolumn{2}{c|}{5$\times$128} & \multicolumn{2}{c}{5$\times$256} \\ 
 & Forward & Backward & Forward & Backward & Forward & Backward & Forward & Backward
\begin{filecontents*}{figures/ied_data/solve_symmetric_stopFalse_float32/time_cpu.csv}
method,forward,backward,forward,backward,forward,backward,forward,backward,forward,backward,forward,backward
AutoDiff-unroll,0.0004,0.0001,0.0007,0.0002,0.0022,0.0003,0.0075,0.0011,-,-,-,-
AutoDiff-DDN-E,0.0004,0.0491,0.0008,0.0942,0.0021,0.1909,0.0080,2.0692,-,-,-,-
PI-unroll,0.0047,0.0121,0.0049,0.0175,0.0051,0.0344,0.0058,0.0984,-,-,-,-
PI-DDN-J,0.0038,0.0014,0.0039,0.0026,0.0042,0.0096,0.0049,0.0462,-,-,-,-
PI-IFT-J,0.0037,0.0230,0.0039,0.0452,0.0041,0.1103,0.0052,1.0283,-,-,-,-
\rowcolor{lightgray} PI-DDN-E,0.0037,0.0005,0.0039,0.0007,0.0041,0.0012,0.0048,0.0025,-,-,-,-
\rowcolor{lightgray} PI-IFT-E,0.0038,0.0003,0.0040,0.0004,0.0041,0.0007,0.0056,0.0018,-,-,-,-
SI-unroll,0.0138,0.0133,0.0401,0.0206,0.1198,0.0437,0.3927,0.1322,-,-,-,-
SI-DDN-J,0.0127,0.0014,0.0387,0.0026,0.1159,0.0096,0.3552,0.0464,-,-,-,-
SI-IFT-J,0.0127,0.0230,0.0383,0.0452,0.1096,0.1098,0.3791,1.0431,-,-,-,-
\rowcolor{lightgray} SI-DDN-E,0.0127,0.0005,0.0384,0.0007,0.1131,0.0012,0.3666,0.0028,-,-,-,-
\rowcolor{lightgray} SI-IFT-E,0.0127,0.0003,0.0393,0.0004,0.1139,0.0007,0.3667,0.0015,-,-,-,-
\end{filecontents*}
\csvreader[head to column names]{figures/ied_data/solve_symmetric_stopFalse_float32/time_cpu.csv}{}
{\\\hline\csvcoli & \csvcolii & \csvcoliii & \csvcoliv & \csvcolv & \csvcolvi & \csvcolvii & \csvcolviii & \csvcolix}
\\\hline
\end{tabular}
}
\end{table*}

\begin{table*}[!ht]
\centering
\small
\caption{\textit{Implicit eigen decomposition (IED) layer, GPU time (s) for symmetric $\mb{A}$.
``AutoDiff": PyTorch \texttt{eigh}() function;
``DDN": deep declarative network;
``IFT": implicit function theorem;
``unroll": unrolling the forward iteration for gradients via PyTorch autodiff mechanism;
``J": autodiff Jacobian without exploited structure;
``E": our implicit differentiation with exploited structure.
Our \colorbox{gray!50}{best suggestions} are highlighted considering the overall solution precision and computational requirements.}
}
\label{tb:ied_symmetric_time_gpu}
\setlength{\tabcolsep}{2pt}
\resizebox{\textwidth}{!}{
\begin{tabular}{l|rr|rr|rr|rr|rr|rr}
\hline
\multicolumn{1}{c|}{\multirow{2}{*}{Method}} & \multicolumn{2}{c|}{5$\times$32} & \multicolumn{2}{c|}{5$\times$64} & \multicolumn{2}{c|}{5$\times$128} & \multicolumn{2}{c|}{5$\times$256} & \multicolumn{2}{c|}{5$\times$512} & \multicolumn{2}{c}{5$\times$1024} \\ 
 & Forward & Backward & Forward & Backward & Forward & Backward & Forward & Backward & Forward & Backward & Forward & Backward
\begin{filecontents*}{figures/ied_data/solve_symmetric_stopFalse_float32/time_gpu.csv}
method,forward,backward,forward,backward,forward,backward,forward,backward,forward,backward,forward,backward
AutoDiff-unroll,0.0004,0.0003,0.0041,0.0003,0.0097,0.0003,0.0261,0.0003,0.0758,0.0005,0.2193,0.0008
AutoDiff-DDN-E,0.0004,0.0784,0.0042,0.1523,0.0102,0.2946,0.0274,0.5861,0.0765,1.1695,0.2258,2.3861
PI-unroll,0.0077,0.0168,0.0077,0.0168,0.0079,0.0171,0.0079,0.0174,0.0082,0.0172,0.0086,0.0174
PI-DDN-J,0.0065,0.0028,0.0066,0.0050,0.0068,0.0096,0.0067,0.0182,0.0069,0.0436,0.0068,0.5281
PI-IFT-J,0.0065,0.0427,0.0065,0.0840,0.0067,0.1685,0.0068,0.3313,0.0070,0.6942,-,-
\rowcolor{lightgray} PI-DDN-E,0.0065,0.0065,0.0065,0.0067,0.0067,0.0033,0.0068,0.0059,0.0069,0.0114,0.0070,0.0414
\rowcolor{lightgray} PI-IFT-E,0.0065,0.0007,0.0066,0.0009,0.0068,0.0013,0.0068,0.0023,0.0069,0.0043,0.0070,0.0152
SI-unroll,0.0591,0.0173,0.1831,0.0169,0.2679,0.0227,0.4631,0.0580,1.0811,0.1443,2.6506,0.4876
SI-DDN-J,0.0590,0.0029,0.1847,0.0051,0.2694,0.0089,0.4642,0.0173,1.0623,0.0419,2.5715,0.1566
SI-IFT-J,0.0588,0.0426,0.1833,0.0835,0.2677,0.1663,0.4642,0.3380,1.0624,0.7120,-,-
\rowcolor{lightgray} SI-DDN-E,0.0584,0.0045,0.1832,0.0021,0.2676,0.0032,0.4646,0.0058,1.0622,0.0113,2.5676,0.0396
\rowcolor{lightgray} SI-IFT-E,0.0583,0.0007,0.1833,0.0009,0.2676,0.0013,0.4641,0.0022,1.0625,0.0042,2.6315,0.0144
\end{filecontents*}
\csvreader[head to column names]{figures/ied_data/solve_symmetric_stopFalse_float32/time_gpu.csv}{}
{\\\hline\csvcoli & \csvcolii & \csvcoliii & \csvcoliv & \csvcolv & \csvcolvi & \csvcolvii & \csvcolviii & \csvcolix & \csvcolx & \csvcolxi & \csvcolxii & \csvcolxiii}
\\\hline
\end{tabular}
}
\end{table*}

\begin{table*}[!ht]
\centering
\small
\caption{\textit{Implicit eigen decomposition (IED) layer, GPU memory (MB) for symmetric $\mb{A}$.
``AutoDiff": PyTorch \texttt{eigh}() function;
``DDN": deep declarative network;
``IFT": implicit function theorem;
``unroll": unrolling the forward iteration for gradients via PyTorch autodiff mechanism;
``J": autodiff Jacobian without exploited structure;
``E": our implicit differentiation with exploited structure.
Our \colorbox{gray!50}{best suggestions} are highlighted considering the overall solution precision and computational requirements.}
}
\label{tb:ied_symmetric_memory}
\setlength{\tabcolsep}{0.7pt}
\resizebox{\textwidth}{!}{
\begin{tabular}{l|rr|rr|rr|rr|rr|rr}
\hline
\multicolumn{1}{c|}{\multirow{2}{*}{Method}} & \multicolumn{2}{c|}{5$\times$32} & \multicolumn{2}{c|}{5$\times$64} & \multicolumn{2}{c|}{5$\times$128} & \multicolumn{2}{c|}{5$\times$256} & \multicolumn{2}{c|}{5$\times$512} & \multicolumn{2}{c}{5$\times$1024} \\ 
 & Forward & Backward & Forward & Backward & Forward & Backward & Forward & Backward & Forward & Backward & Forward & Backward
\begin{filecontents*}{figures/ied_data/solve_symmetric_stopFalse_float32/memory_gpu.csv}
method,forward,backward,forward,backward,forward,backward,forward,backward,forward,backward,forward,backward
AutoDiff-unroll,0.5903,0.0972,0.8247,0.3896,1.7671,1.5605,5.5298,6.2456,20.5552,24.9907,80.6060,99.9810
AutoDiff-DDN-E,0.5903,0.0884,0.8232,0.3467,1.7646,1.3789,5.5249,5.5063,20.5454,22.0112,80.5864,88.0210
PI-unroll,0.8228,0.0537,0.9810,0.2280,1.4146,0.9282,2.8506,3.7334,7.5977,14.9688,24.5918,59.9395
PI-DDN-J,0.5293,0.6743,0.5898,5.1694,0.8281,40.6440,1.7759,322.5327,5.5464,2570.0620,20.5874,20520.1206
PI-IFT-J,0.5293,1.2705,0.5898,10.0791,0.8281,80.3135,1.7759,641.2510,5.5464,5125.0010,-,-
\rowcolor{lightgray} PI-DDN-E,0.5293,0.0674,0.5898,0.2534,0.8281,1.0044,1.7759,4.0068,5.5464,16.0117,20.5874,64.0215
\rowcolor{lightgray} PI-IFT-E,0.5293,0.1230,0.5898,0.4868,0.8281,1.9409,1.7759,7.7559,5.5464,31.0107,20.5874,124.0205
SI-unroll,4.6211,0.0962,16.5869,0.3882,65.0103,1.5576,256.5151,6.2402,1022.0249,24.9805,4083.0444,99.9609
SI-DDN-J,0.7539,0.6743,1.1182,5.1694,3.1353,40.6440,9.0151,322.5327,32.0249,2570.0620,123.0444,20520.1206
SI-IFT-J,0.7539,1.2705,1.1182,10.0791,3.1353,80.3135,9.0151,641.2510,32.0249,5125.0010,-,-
\rowcolor{lightgray} SI-DDN-E,0.7539,0.0674,1.1182,0.2534,3.1353,1.0044,9.0151,4.0068,32.0249,16.0117,123.0444,64.0215
\rowcolor{lightgray} SI-IFT-E,0.7539,0.1230,1.1182,0.4868,3.1353,1.9409,9.0151,7.7559,32.0249,31.0107,123.0444,124.0205
\end{filecontents*}
\csvreader[head to column names]{figures/ied_data/solve_symmetric_stopFalse_float32/memory_gpu.csv}{}
{\\\hline\csvcoli & \csvcolii & \csvcoliii & \csvcoliv & \csvcolv & \csvcolvi & \csvcolvii & \csvcolviii & \csvcolix & \csvcolx & \csvcolxi & \csvcolxii & \csvcolxiii}
\\\hline
\end{tabular}
}
\end{table*}

\begin{table*}[!ht]
\centering
\small
\caption{\textit{Implicit eigen decomposition (IED) layer, CPU time (s) for nonsymmetric $\mb{A}$.
``AutoDiff": PyTorch \texttt{eigh}() function;
``DDN": deep declarative network;
``IFT": implicit function theorem;
``unroll": unrolling the forward iteration for gradients via PyTorch autodiff mechanism;
``J": autodiff Jacobian without exploited structure;
``E": our implicit differentiation with exploited structure.
Our \colorbox{gray!50}{best suggestions} are highlighted considering the overall solution precision and computational requirements.}
}
\label{tb:ied_nonsymmetric_time_cpu}
\setlength{\tabcolsep}{11pt}
\resizebox{\textwidth}{!}{
\begin{tabular}{l|rr|rr|rr|rr}
\hline
\multicolumn{1}{c|}{\multirow{2}{*}{Method}} & \multicolumn{2}{c|}{5$\times$32} & \multicolumn{2}{c|}{5$\times$64} & \multicolumn{2}{c|}{5$\times$128} & \multicolumn{2}{c}{5$\times$256} \\ 
 & Forward & Backward & Forward & Backward & Forward & Backward & Forward & Backward
\begin{filecontents*}{figures/ied_data/solve_nonsymmetric_stopFalse_float32/time_cpu.csv}
method,forward,backward,forward,backward,forward,backward,forward,backward,forward,backward,forward,backward
AutoDiff-unroll,0.0004,0.0001,0.0008,0.0002,0.0022,0.0003,0.0075,0.0011,-,-,-,-
AutoDiff-DDN-E,0.0004,0.0483,0.0008,0.0925,0.0021,0.1902,0.0075,2.0929,-,-,-,-
PI-unroll,0.0045,0.0113,0.0046,0.0150,0.0048,0.0306,0.0054,0.0923,-,-,-,-
PI-DDN-J,0.0036,0.0014,0.0037,0.0026,0.0039,0.0096,0.0045,0.0463,-,-,-,-
PI-IFT-J,0.0037,0.0221,0.0038,0.0432,0.0040,0.1040,0.0050,0.9539,-,-,-,-
\rowcolor{lightgray} PI-DDN-E,0.0036,0.0004,0.0037,0.0006,0.0039,0.0012,0.0044,0.0024,-,-,-,-
\rowcolor{lightgray} PI-IFT-E,0.0036,0.0003,0.0038,0.0004,0.0039,0.0006,0.0046,0.0015,-,-,-,-
SI-unroll,0.0136,0.0131,0.0396,0.0198,0.1159,0.0426,0.3661,0.1298,-,-,-,-
SI-DDN-J,0.0127,0.0014,0.0387,0.0027,0.1182,0.0097,0.3651,0.0468,-,-,-,-
SI-IFT-J,0.0125,0.0220,0.0380,0.0428,0.1104,0.1045,0.3977,1.0048,-,-,-,-
\rowcolor{lightgray} SI-DDN-E,0.0126,0.0004,0.0388,0.0006,0.1162,0.0012,0.3617,0.0028,-,-,-,-
\rowcolor{lightgray} SI-IFT-E,0.0127,0.0003,0.0385,0.0004,0.1184,0.0007,0.3739,0.0016,-,-,-,-
\end{filecontents*}
\csvreader[head to column names]{figures/ied_data/solve_nonsymmetric_stopFalse_float32/time_cpu.csv}{}
{\\\hline\csvcoli & \csvcolii & \csvcoliii & \csvcoliv & \csvcolv & \csvcolvi & \csvcolvii & \csvcolviii & \csvcolix }
\\\hline
\end{tabular}
}
\end{table*}

\begin{table*}[!ht]
\centering
\small
\caption{\textit{Implicit eigen decomposition (IED) layer, GPU time (s) for nonsymmetric $\mb{A}$.
``AutoDiff": PyTorch \texttt{eigh}() function;
``DDN": deep declarative network;
``IFT": implicit function theorem;
``unroll": unrolling the forward iteration for gradients via PyTorch autodiff mechanism;
``J": autodiff Jacobian without exploited structure;
``E": our implicit differentiation with exploited structure.
Our \colorbox{gray!50}{best suggestions} are highlighted considering the overall solution precision and computational requirements.}
}
\label{tb:ied_nonsymmetric_time_gpu}
\setlength{\tabcolsep}{2pt}
\resizebox{\textwidth}{!}{
\begin{tabular}{l|rr|rr|rr|rr|rr|rr}
\hline
\multicolumn{1}{c|}{\multirow{2}{*}{Method}} & \multicolumn{2}{c|}{5$\times$32} & \multicolumn{2}{c|}{5$\times$64} & \multicolumn{2}{c|}{5$\times$128} & \multicolumn{2}{c|}{5$\times$256} & \multicolumn{2}{c|}{5$\times$512} & \multicolumn{2}{c}{5$\times$1024} \\ 
 & Forward & Backward & Forward & Backward & Forward & Backward & Forward & Backward & Forward & Backward & Forward & Backward
\begin{filecontents*}{figures/ied_data/solve_nonsymmetric_stopFalse_float32/time_gpu.csv}
method,forward,backward,forward,backward,forward,backward,forward,backward,forward,backward,forward,backward
AutoDiff-unroll,0.0004,0.0003,0.0041,0.0003,0.0097,0.0003,0.0262,0.0003,0.0759,0.0005,0.2188,0.0008
AutoDiff-DDN-E,0.0004,0.0816,0.0042,0.1586,0.0102,0.3062,0.0275,0.5802,0.0766,1.1564,0.2271,2.3584
PI-unroll,0.0072,0.0166,0.0072,0.0165,0.0074,0.0166,0.0074,0.0166,0.0077,0.0168,0.0080,0.0170
PI-DDN-J,0.0060,0.0028,0.0061,0.0050,0.0063,0.0090,0.0063,0.0174,0.0064,0.0431,0.0063,0.5642
PI-IFT-J,0.0061,0.0411,0.0061,0.0811,0.0065,0.1613,0.0064,0.3218,0.0066,0.6743,-,-
\rowcolor{lightgray} PI-DDN-E,0.0061,0.0064,0.0061,0.0068,0.0069,0.0033,0.0063,0.0059,0.0065,0.0113,0.0067,0.0410
\rowcolor{lightgray} PI-IFT-E,0.0060,0.0006,0.0060,0.0008,0.0064,0.0012,0.0064,0.0021,0.0065,0.0041,0.0066,0.0151
SI-unroll,0.0583,0.0166,0.1830,0.0163,0.2674,0.0226,0.4646,0.0582,1.0884,0.1454,2.6464,0.4886
SI-DDN-J,0.0588,0.0029,0.1847,0.0050,0.2692,0.0089,0.4624,0.0173,1.0584,0.0420,2.5674,0.1567
SI-IFT-J,0.0590,0.0403,0.1832,0.0793,0.2671,0.1581,0.4623,0.3211,1.0582,0.6744,-,-
\rowcolor{lightgray} SI-DDN-E,0.0584,0.0045,0.1833,0.0021,0.2671,0.0031,0.4625,0.0057,1.0583,0.0112,2.5622,0.0395
\rowcolor{lightgray} SI-IFT-E,0.0583,0.0007,0.1833,0.0009,0.2671,0.0013,0.4624,0.0022,1.0590,0.0041,2.6394,0.0144
\end{filecontents*}
\csvreader[head to column names]{figures/ied_data/solve_nonsymmetric_stopFalse_float32/time_gpu.csv}{}
{\\\hline\csvcoli & \csvcolii & \csvcoliii & \csvcoliv & \csvcolv & \csvcolvi & \csvcolvii & \csvcolviii & \csvcolix & \csvcolx & \csvcolxi & \csvcolxii & \csvcolxiii}
\\\hline
\end{tabular}
}
\end{table*}

\begin{table*}[!ht]
\centering
\small
\caption{\textit{Implicit eigen decomposition (IED) layer, GPU memory (MB) for nonsymmetric $\mb{A}$.
``AutoDiff": PyTorch \texttt{eigh}() function;
``DDN": deep declarative network;
``IFT": implicit function theorem;
``unroll": unrolling the forward iteration for gradients via PyTorch autodiff mechanism;
``J": autodiff Jacobian without exploited structure;
``E": our implicit differentiation with exploited structure.
Our \colorbox{gray!50}{best suggestions} are highlighted considering the overall solution precision and computational requirements.}
}
\label{tb:ied_nonsymmetric_memory}
\setlength{\tabcolsep}{0.7pt}
\resizebox{\textwidth}{!}{
\begin{tabular}{l|rr|rr|rr|rr|rr|rr}
\hline
\multicolumn{1}{c|}{\multirow{2}{*}{Method}} & \multicolumn{2}{c|}{5$\times$32} & \multicolumn{2}{c|}{5$\times$64} & \multicolumn{2}{c|}{5$\times$128} & \multicolumn{2}{c|}{5$\times$256} & \multicolumn{2}{c|}{5$\times$512} & \multicolumn{2}{c}{5$\times$1024} \\ 
 & Forward & Backward & Forward & Backward & Forward & Backward & Forward & Backward & Forward & Backward & Forward & Backward
\begin{filecontents*}{figures/ied_data/solve_nonsymmetric_stopFalse_float32/memory_gpu.csv}
method,forward,backward,forward,backward,forward,backward,forward,backward,forward,backward,forward,backward
AutoDiff-unroll,0.5747,0.0972,0.8091,0.3896,1.7515,1.5605,5.5142,6.2456,20.5396,24.9907,80.5903,99.9810
AutoDiff-DDN-E,0.5747,0.0884,0.8076,0.3467,1.7490,1.3789,5.5093,5.5063,20.5298,22.0112,80.5708,88.0210
PI-unroll,0.8071,0.0537,0.9653,0.2280,1.3989,0.9282,2.8350,3.7334,7.5820,14.9688,24.5762,59.9395
PI-DDN-J,0.5137,0.6743,0.5742,5.1694,0.8125,40.6440,1.7603,322.5327,5.5308,2570.0620,20.5718,20520.1206
PI-IFT-J,0.5137,1.2705,0.5742,10.0791,0.8125,80.3135,1.7603,641.2510,5.5308,5125.0010,-,-
\rowcolor{lightgray} PI-DDN-E,0.5137,0.0674,0.5742,0.2534,0.8125,1.0044,1.7603,4.0068,5.5308,16.0117,20.5718,64.0215
\rowcolor{lightgray} PI-IFT-E,0.5137,0.1230,0.5742,0.4868,0.8125,1.9409,1.7603,7.7559,5.5308,31.0107,20.5718,124.0205
SI-unroll,4.6055,0.0962,16.5713,0.3882,64.9946,1.5576,256.4995,6.2402,1022.0093,24.9805,4083.0288,99.9609
SI-DDN-J,0.7383,0.6743,1.1025,5.1694,3.1196,40.6440,8.9995,322.5327,32.0093,2570.0620,123.0288,20520.1206
SI-IFT-J,0.7383,1.2705,1.1025,10.0791,3.1196,80.3135,8.9995,641.2510,32.0093,5125.0010,-,-
\rowcolor{lightgray} SI-DDN-E,0.7383,0.0674,1.1025,0.2534,3.1196,1.0044,8.9995,4.0068,32.0093,16.0117,123.0288,64.0215
\rowcolor{lightgray} SI-IFT-E,0.7383,0.1230,1.1025,0.4868,3.1196,1.9409,8.9995,7.7559,32.0093,31.0107,123.0288,124.0205
\end{filecontents*}
\csvreader[head to column names]{figures/ied_data/solve_nonsymmetric_stopFalse_float32/memory_gpu.csv}{}
{\\\hline\csvcoli & \csvcolii & \csvcoliii & \csvcoliv & \csvcolv & \csvcolvi & \csvcolvii & \csvcolviii & \csvcolix & \csvcolx & \csvcolxi & \csvcolxii & \csvcolxiii}
\\\hline
\end{tabular}
}
\end{table*}